\documentclass[10pt,twocolumn,letterpaper]{article}

\usepackage[pagenumbers]{cvpr} 

\definecolor{cvprblue}{rgb}{0.21,0.49,0.74}
\usepackage[pagebackref,breaklinks,colorlinks,allcolors=cvprblue]{hyperref}
\usepackage[utf8]{inputenc} 
\usepackage[T1]{fontenc}    
\usepackage{url}            
\usepackage{booktabs}       
\usepackage{amsfonts}       
\usepackage{nicefrac}       
\usepackage{microtype}      
\usepackage[table]{xcolor}         
\usepackage{graphicx}
\usepackage{float}
\usepackage{caption}
\usepackage{multirow}
\usepackage{fancyhdr}
\usepackage{enumitem}
\usepackage{amsthm}
\usepackage{amsmath}
\usepackage{amssymb}
\usepackage{mathtools}
\usepackage{algorithm}
\usepackage{algorithmic}
\usepackage{wrapfig}
\usepackage{lipsum}
\usepackage[most]{tcolorbox}
\usepackage{fancybox,framed}
\usepackage{threeparttable} 
\usepackage{mdframed}
\usepackage{fontawesome}
\usepackage{pgffor}
\usepackage{bbm}
\definecolor{SelfColor}{rgb}{0.913,0.443,0.196}
\definecolor{UrlColor}{rgb}{0.776, 0.255, 0.275}
\definecolor{RefColor}{rgb}{0.290,0.565,0.596}
\definecolor{TableColor}{rgb}{0.737,0.741,0.275}

\definecolor{mydarkgreen}{rgb}{0.267,0.482,0.235}
\definecolor{mydarkred}{rgb}{0.714,0.220,0.192}

\newcommand{\up}{%
    \rlap{\hspace{-0.1pt}{%
        \scriptsize\textcolor{mydarkgreen}{\textsuperscript{\textbf{+}}}}}%
}

\newcommand{\dn}{%
    \rlap{\hspace{-0.1pt}{%
        \scriptsize\textcolor{mydarkred}{\textsuperscript{\textbf{--}}}}}%
}

\newcommand{\sa}{%
    \rlap{\hspace{-0.1pt}{%
        \scriptsize\textcolor{gray}{\textsuperscript{\textbf{±}}}}}%
}

\def\confName{CVPR}
\def\confYear{2027}

\title{Beyond Attribute-Centric Fairness: Modelling and Mitigating Unified Bias Representation in Omni-modal Large Language Models}

\author{Runqi Lin$^{1}$ \, Suqin Yuan$^{2}$ \, Zihao Fu$^{3}$ \, Chris Russell$^{1}$ \\
$^1$University of Oxford \,
$^2$The University of Sydney\,
$^3$The Chinese University of Hong Kong\\
}

\begin{document}
\maketitle

\begingroup
\renewcommand{\thefootnote}{}
\footnotetext[0]{A Preprint.}
\endgroup

\begin{abstract}

Omni-modal large language models (OLLMs) provide joint multimodal understanding and generation capabilities, making them well-suited to complex environments and promising backbones for next-generation AI.
Despite these advances, the harmful stereotypes encoded in such models manifest in increasingly diverse forms, leading to broader and more severe societal consequences.
Existing debiasing methods typically impose separate fairness objectives for each sensitive attribute and modality to minimise demographic-dependent variations.
However, in omni-modal settings, this attribute-centric paradigm struggles to jointly address diverse social biases with distinct fairness objectives and to generalise beyond predefined attributes.
To this end, we propose OmniDebias, one of the first methods for promoting fairness in OLLMs, which employs a bias-centric formulation to extract and mitigate a shared bias representation across attributes and modalities while preserving attribute-related knowledge.
Specifically, we exploit activation inconsistencies of a given attribute (e.g., gender) across semantic contexts (e.g., occupations) to disentangle bias from legitimate knowledge.
We further consolidate multiple extracted biases into a shared representation, formulating a unified objective that captures the underlying structure across social biases and generalises to unseen attributes.
Extensive experiments demonstrate that eliminating unified bias with both post-hoc editing and fine-tuning can consistently improve fairness across diverse attributes and modalities in OLLMs while maintaining favourable utility.


\end{abstract}
    
\section{Introduction}

To handle the diverse demands of complex scenarios, omni-modal large language models (OLLMs) utilise a joint embedding space to process cross-modal information, offering a unified paradigm for multimodal understanding and generation~\citep{OpenAI, Claude, Gemini}.
Despite their remarkable performance and great potential, the increasing deployment of OLLMs in real-world systems has raised societal concerns about their potential risks~\citep{yang2025reliable, xu2025mmdt}. 
Recent studies suggest that their multimodal nature leads to more diverse manifestations of harmful stereotypes, which in turn may propagate social biases and imperceptibly shape public perception in unintended ways~\citep{gallegos2024bias, adewumi2024fairness, zhang2025debiasing, elobaid2026demographic}.

To mitigate such biases, previous works have attempted to incorporate fairness constraints throughout the model lifecycle.
Prompt-based methods modify the input instructions to either explicitly assign a particular attribute class to each input or provide guidance to encourage equal opportunity across all classes~\citep{bansal2022well, sakurai2025fairt2i}.
Fine-tuning-based (FT) methods equalise model predictions with respect to targeted attributes by training on a demographically balanced dataset or optimising fairness-aware objectives~\citep{zhang2023iti, li2025fair}.
Post-hoc editing (PE) methods steer model outputs at inference time by sampling an attribute class randomly and injecting its representation into the model's activations~\citep{park2025fair, fu2026fairimagen}.

\begin{figure*}[t]
\centering    
\includegraphics[width=1.9\columnwidth]{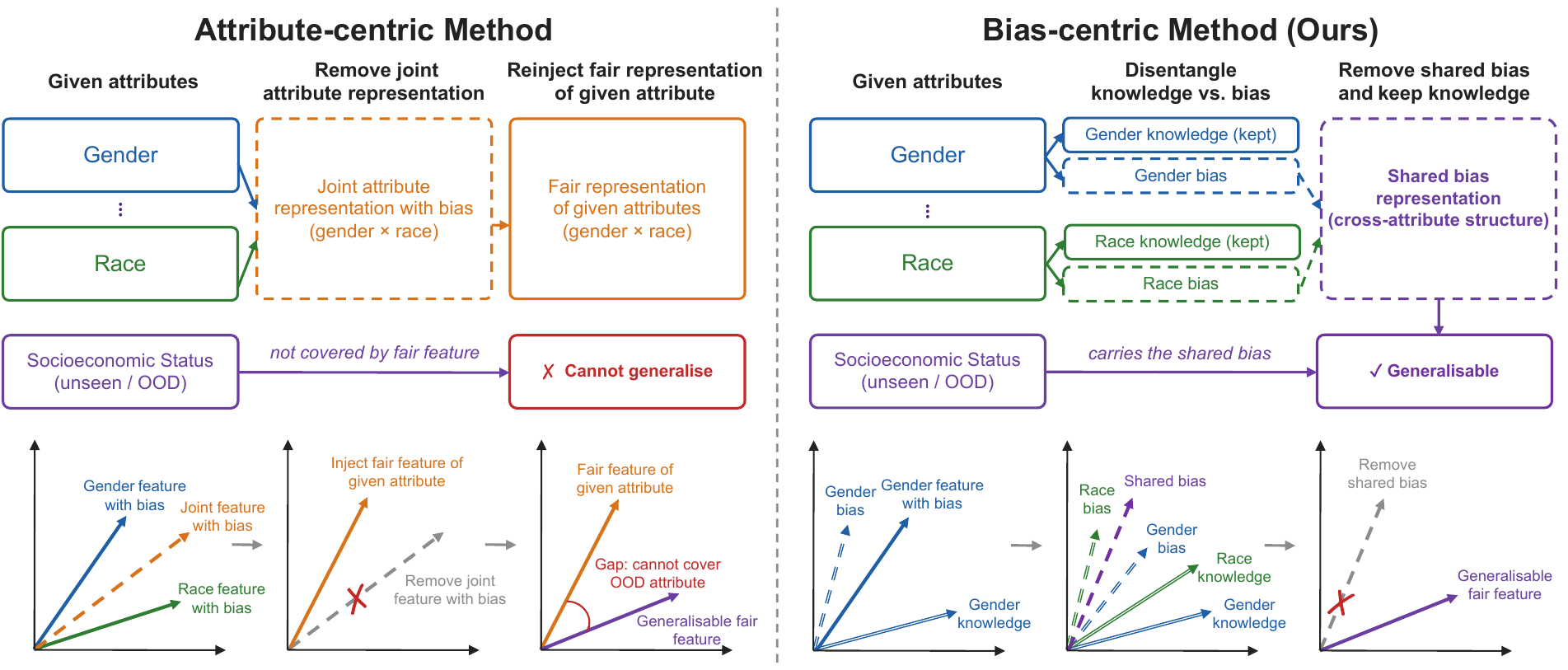}   
\vspace{-.5em}
\caption{
Schematic comparison of attribute-centric (left panel) and bias-centric (right panel) debiasing paradigms, showing their workflows (top row) and feature-space geometry (bottom row). 
Attribute-centric methods replace bias-contaminated directions with predefined unbiased ones, requiring target-specific priors that limit generalisation to unseen attributes. Bias-centric methods instead disentangle and mitigate shared bias while preserving legitimate attribute-related knowledge, enabling unified debiasing across attributes and modalities and facilitating generalisation to unseen attributes.
}
\label{fig:draw}
\vspace{-1.5em}
\end{figure*}

However, all the aforementioned methods are inherently attribute-centric, requiring separate fairness constraints for each predefined attribute and modality~\citep{li2024self, shenfinetuning}.
As more attributes and modalities are considered, attribute-centric methods need to accommodate multiple fairness objectives with distinct targets within the same space, inducing mutual interference and becoming increasingly difficult to satisfy simultaneously.
Furthermore, the demographic concepts and their interactions in OLLMs are infeasible to enumerate, inevitably posing an out-of-distribution (OOD) challenge that current paradigms struggle to handle in the absence of prescribed supervision.
Consequently, as shown in Figure~\ref{fig:draw}, attribute-centric methods fall short of providing scalable fairness for OLLMs, in which social biases manifest in diverse and intricately entangled ways.

To overcome these limitations, we propose OmniDebias, among the first methods to enable fairness interventions across diverse social concepts in OLLMs.
It adopts a bias-centric formulation that extracts shared bias across attributes and modalities and uses it as a joint fairness target, enabling unified debiasing without separate attribute-specific objectives.
Specifically, we first disentangle bias from the legitimate knowledge associated with each attribute by exploiting activation inconsistencies across semantic contexts.
Ideally, switching an attribute class should induce the same activation shift across different contexts, as the transformation itself is context-invariant; e.g., the activation differences between $\text{male CEO}\leftrightarrow\text{female CEO}$ and $\text{male nurse}\leftrightarrow\text{female nurse}$ should be consistent.
However, in practice, these attribute-induced shifts are inconsistent across contexts due to the undesirable entanglement of stereotypes with legitimate knowledge.
We therefore exploit these deviations to isolate and mitigate the entangled bias within each attribute while preserving the attribute-related knowledge.
Although legitimate knowledge varies across attributes, the extracted biases arise from social stereotypes and may exhibit shared characteristics across attributes.
We thus consolidate the extracted biases into a unified representation, accentuating shared bias patterns to better characterise the underlying stereotypes in OLLMs and facilitate generalisation to OOD attributes.

Building on the extracted shared bias, we establish a unified fairness target for mitigating social stereotypes across multiple attributes and modalities in OLLMs.
This bias-centric target can be readily instantiated in both post-hoc editing (PE) and fine-tuning (FT), without specifying separate fair representations for individual attributes. 
For PE, we mitigate the entangled social bias by projecting the model parameters onto its orthogonal direction at inference time~\citep{arditi2024refusal,fu2026fairimagen}.
For FT, we dynamically estimate the residual bias in the evolving model and penalise it through a regularisation term.
Notably, the bias estimate is derived solely from input activations, without requiring ground-truth labels or additional supervision.
In both settings, OmniDebias consistently improves fairness for seen and OOD sensitive attributes across all three modalities while preserving generation quality. Our main contributions are summarised as follows:
\begin{itemize}[leftmargin=10pt, topsep=0pt, itemsep=2pt]

\item We introduce a bias-centric formulation that disentangles bias from a given sensitive attribute while preserving the corresponding legitimate knowledge.

\item We establish a unified fairness objective by consolidating biases across diverse attributes and modalities, enabling scalable fairness and generalisation to unseen attributes.

\item Extensive experiments demonstrate that OmniDebias consistently improves fairness while maintaining favourable model utility under both fine-tuning and post-hoc editing.

\end{itemize}

\section{Related Work}


\textbf{Omni-modal Large Language Models.}\hspace*{2mm}Representative OLLMs, such as Janus~\citep{chen2025janus} and BAGEL~\citep{deng2025emerging}, typically adopt multi-branch architectures that fuse modality-specific representations into a shared reasoning space, supporting both multimodal understanding and generation.
A common design principle is to use modality-specific encoders and projectors to independently encode raw inputs and map them into a shared representation space, where the resulting features are jointly processed by a shared self-attention backbone before being decoded by task-specific understanding or generation heads.


\textbf{Prompt-based Method.}\hspace*{2mm}A straightforward strategy for mitigating social bias is to augment input instructions with fairness-aware guidance.
In both text-to-text (T2T) and text-to-image (T2I) generation, prompts are typically augmented with explicit demographic classes~\citep{bansal2022well, li2025prompting}, or guided by manually crafted or LLM-generated ethical interventions~\citep{anantaprayoon2025intent, sakurai2025fairt2i}.
For image-to-text (I2T) description, multimodal prompts are employed to mitigate demographic biases by encouraging more equitable interpretations of visual content~\citep{wu2025evaluating}.
Despite their simplicity, such methods still require a separate directive for predefined attributes and modalities and provide limited guarantees of fairness in the resulting outputs.

\textbf{Post-hoc Editing Method.}\hspace*{2mm}Another line of fairness research explores inference-time interventions that randomly sample an attribute class and steer model outputs along its corresponding direction to promote balanced outcomes.
In T2T settings, these methods typically identify stereotypical associations through self-debiasing and mitigate their influence on model predictions via reprompting or logit calibration~\citep{bolukbasi2016man, ravfogel2020null, gallegos2025self, di2024post}.
The T2I PE methods are often performed directly on prompt embeddings, for example, by projecting them into a fair subspace to suppress demographic information while preserving semantic content~\citep{fu2026fairimagen, park2025fair}.
For I2T fairness, these methods mitigate bias-associated residuals or selectively modulate demographic neurons to reduce biased responses~\citep{lan2025my, an2026interpretable}.
However, these methods still require attribute-specific directions and cannot generalise to unseen ones, while compressing multiple demographic representations into a shared space also inevitably loses information, making them ill-suited to OLLMs.

\textbf{Fine-tuning-based Method.}\hspace*{2mm}FT fairness methods directly update model parameters to suppress social bias. 
The T2T fairness methods typically regularise bias directions or intermediate layers with fairness losses~\citep{li2024mitigating, zhang2024bias}. 
To improve efficiency, some work utilises lightweight modules to provide calibration or reasoning guidance without full retraining~\citep{chen2025identifying, kabra2025reasoning}.
For T2I generation, fairness is often formulated as distribution alignment or encoder-level control, using fairness-aware mapping layers to steer demographic distributions~\citep{shenfinetuning, li2025fair}.
The I2T fairness methods usually minimise mutual information of attributes to learn demographic-invariant representations~\citep{bhosale2026fairllava} or align multimodal representations to reduce stereotypical associations in image-conditioned outputs~\citep{cheng2025social}.
Nevertheless, these methods remain attribute-centric and require ground-truth supervision, limiting their scalability to the diverse and complex social concepts in OLLMs.

\section{Methodology}

In this section, we introduce OmniDebias, a bias-centric framework that disentangles bias from legitimate knowledge across semantic contexts (Section~\ref{section:3_1}), unifies the extracted biases across attributes and modalities into a shared representation (Section~\ref{section:3_2}), and mitigates the resulting bias through both PE and FT to consistently promote fairness while preserving attribute-related knowledge (Section~\ref{section:3_3}).

\subsection{Bias Disentanglement}
\label{section:3_1}

To investigate social stereotypes in OLLMs, we examine the activations associated with sensitive attributes across diverse semantic contexts.
For an input of length $T$ from modality $m\in\mathcal{M}=\{\text{text}, \text{image}, \ldots\}$, a modality-specific tokenizer maps it to an embedding sequence in $\mathbb{R}^{T\times d}$, which is then fed into the shared $L$-layer OLLM backbone $f$\footnote{$L$ denotes the number of modality-shared layers without dedicated components, e.g., T2I flow modules.}.
For each modality $m$, we consider a sensitive attribute $a\in\mathcal{A}_m=\{\text{gender}, \text{race}, \text{age}, \ldots\}$ with attribute classes $\mathcal{S}_{m, a}=\{s_1,s_2,\ldots\}$, such as $\mathcal{S}_{\text{text},\text{gender}}=\{\text{male},\text{female}\}$, and a semantic context $c\in\mathcal{C}=\{\text{CEO}, \text{doctor}, \text{nurse}, \ldots\}$.

We then construct a matched input pair $\big((s_1,c),(s_2,c)\big)$ that shares the same modality, sensitive attribute, and semantic context while differing only in the attribute class, e.g., $\text{male CEO}\leftrightarrow\text{female CEO}$ ($m=\text{text}$).
For attributes with more than two classes, we construct such matched pairs for all pairwise combinations.
For a given pair $s_1,s_2\in\mathcal{S}_{m, a}$, we define the attribute-induced activation difference for attribute $a$ in modality $m$ under context $c$ at layer $l$ as:
\begin{equation}
\Delta_{m, a}^{(l)}(c)
=
h_{m, a}^{(l)}(s_1,c)-h_{m, a}^{(l)}(s_2,c)
\in\mathbb{R}^{d}.
\label{eq:contrast}
\end{equation}
The massive activations in OLLMs~\citep{sun2024massive} may dominate direction estimation and distort the attribute-induced activation differences.
We therefore apply $\alpha$-strength standardisation to the activations based on their coordinate-wise standard deviation $\boldsymbol{\sigma}^{(l)}$, as $h^{(l)} \leftarrow h^{(l)} \oslash (\boldsymbol{\sigma}^{(l)})^{\alpha}$ for all layers $l$, to better estimate the attribute directions\footnote{At the debiasing stage, we rescale the extracted bias directions to their coordinates in the native activation space.}.

\begin{figure*}[t]
    \centering
    \captionsetup{skip=6pt, belowskip=0pt}
    \captionsetup[subfigure]{skip=4pt, belowskip=0pt}

    \newlength{\DemoPanelHeight}
    \setlength{\DemoPanelHeight}{0.42\textwidth}

    \begin{subfigure}[t]{0.45\textwidth}
        \centering
        \includegraphics[height=\DemoPanelHeight]
            {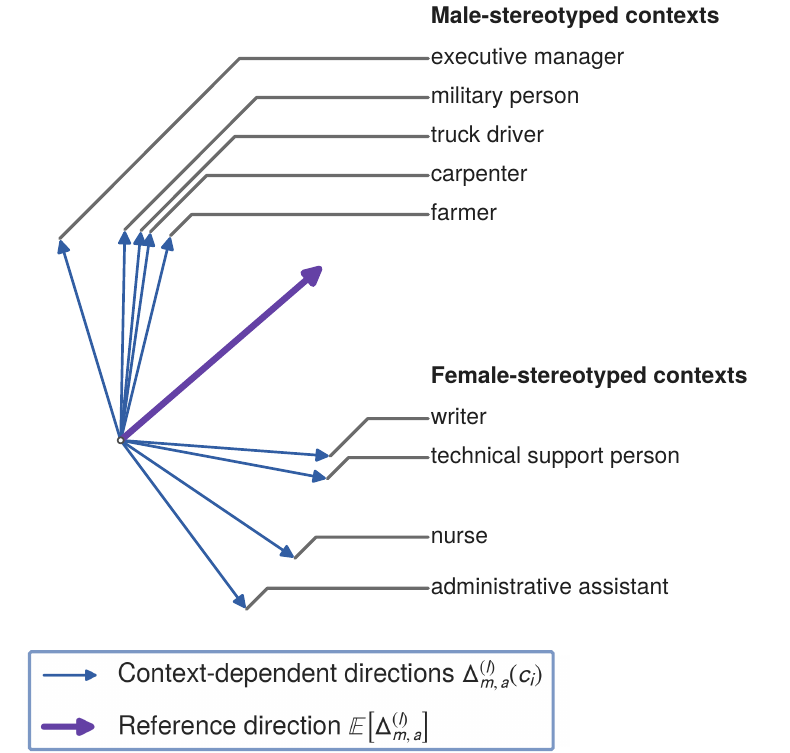}
            \vspace{0.25em}
        \caption{(a) Context-dependent directions
        (Gender $\cdot$ T2T).}
        \label{fig:demo-directions}
    \end{subfigure}\hfill%
    \begin{subfigure}[t]{0.51\textwidth}
        \centering
        \includegraphics[height=\DemoPanelHeight]
            {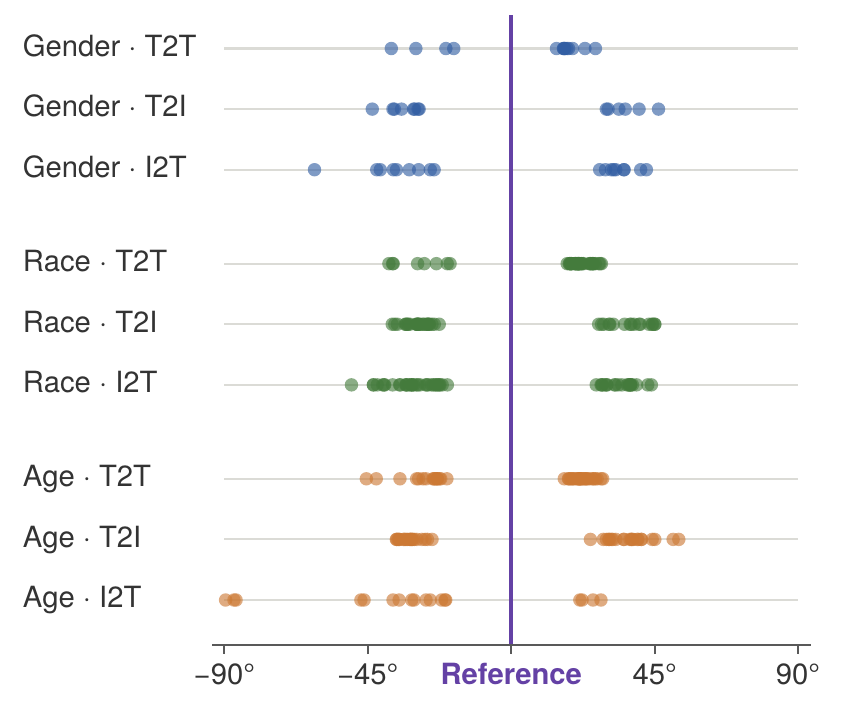}
            
        \vspace{0.15em}
        \caption{(b) Angles between context-dependent
        and reference directions.}
        \label{fig:demo-angles}
    \end{subfigure}

    \caption{Context-dependent activation directions
    (a) and their angular differences relative to the
    reference direction (b) in Janus.
    The gender-related activation directions exhibit
    substantial and structured variations across male-
    and female-stereotyped contexts.
    The angular differences from the reference direction
    across attributes and modalities further reveal
    widespread context-dependent variations.}
    \label{fig:demo}
    \vspace{-1.5em}
\end{figure*}

To mitigate the bias manifested in $\Delta_{m,a}^{(l)}(c)$, existing attribute-centric methods typically introduce a predefined fairness direction for each target attribute and use it to override the contaminated directions across semantic contexts~\citep{fu2026fairimagen,webster2020measuring}.
However, such replacement requires external legitimate knowledge specific to the attribute and modality, without leveraging the legitimate attribute representations already encoded in the model.
Moreover, each unbiased direction encodes distinct attribute-related knowledge, causing mutual interference in multi-target debiasing and offering limited generalisability beyond predefined attributes, as illustrated in Figure~\ref{fig:draw} (left panel).

In contrast, we aim to disentangle bias directly from the contaminated directions across different semantic contexts, while preserving the legitimate knowledge inherently encoded in OLLMs without external priors.
Ideally, the activation difference of attribute pairs should be consistent across semantic contexts, as modality- and context-specific effects cancel each other out and only the intrinsic attribute direction remains: $\Delta_{m,a}^{(l)}(c_i) \approx \Delta_{m,a}^{(l)}(c_j)$.
However, as shown in Figure~\ref{fig:demo}(a), the activation directions exhibit substantial and structured inconsistencies across semantic contexts, with male- and female-stereotyped occupations tending towards opposing orientations.
Moreover, we observe similar patterns across three attributes and three modalities, with deviations from the reference direction averaging approximately $30^\circ$, as illustrated in Figure~\ref{fig:demo}(b).
We attribute these systematic departures to stereotypical patterns encoded in the model, which entangle context-dependent bias with legitimate attribute knowledge.
Therefore, we first follow the common practice of averaging activation differences across semantic contexts to obtain a reference direction $\mathbb{E}[\Delta_{m,a}^{(l)}]$~\citep{fu2026fairimagen,li2025fair}, and then exploit deviations from this reference to estimate context-dependent bias:
\begin{equation}
\delta_{m,a}^{(l)}(c_i)
=
\Delta_{m,a}^{(l)}(c_i)
-
\mathbb{E}
\left[\Delta_{m,a}^{(l)}\right].
\label{eq:bias_disentanglement}
\end{equation}
Following this formulation, we pool the bias residuals across all attribute-class pairs and semantic contexts for attribute $a$ in modality $m$ into a matrix $\mathbf{R}_{m,a}^{(l)}=\left[\delta_{m,a}^{(l)}(c)\right]$.
Although these residuals capture context-dependent deviations, part of the variation may arise from changes in semantic content itself.
To mitigate semantic interference, we adapt FairPCA~\citep{kleindessner2023efficient} with a reversed objective to suppress the semantic-content subspace, yielding a more precise estimate of the underlying bias.
Specifically, we use raw activations without attribute class contrasts and mean-centre them across contexts to capture context-specific semantic representations:
\begin{equation}
\begin{gathered}
\mu_{m}^{(l)}(c_i)
=
\mathbb{E}_{s\sim\mathcal{S}_{m,a}}
\left[
h_{m,a}^{(l)}(s,c_i)
\right],\\[4pt]
\hat{\mu}_{m}^{(l)}(c_i)
=
\mu_{m}^{(l)}(c_i)
-
\frac{1}{|\mathcal{C}|}
\sum_{c\in\mathcal{C}}
\mu_{m}^{(l)}(c).
\end{gathered}
\label{eq:content}
\end{equation}
We stack the centred semantic representations across contexts into
$\mathbf{F}_{m}^{(l)}=[\hat{\mu}_{m}^{(l)}(c)]$.
Let $\operatorname{SVD}_{x}(\mathbf{F})$ denote the matrix formed by the top-$x$
right singular vectors of $\mathbf{F}$, with $x$ determining the number of
semantic-content directions to suppress.
We then remove these directions from the bias residuals via orthogonal projection:
\begin{equation}
\mathbf{R}_{m,a}^{(l)}
\leftarrow
\mathbf{R}_{m,a}^{(l)}
\left[
\mathbf{I}
-
\operatorname{SVD}_{x}\!\left(\mathbf{F}\right)
\operatorname{SVD}_{x}\!\left(\mathbf{F}\right)^{\top}
\right].
\label{eq:deflate}
\end{equation}
Finally, for attribute $a$ in modality $m$, we extract a $y$-dimensional representation of the dominant stereotypical bias from the refined residual matrix while preserving the model's legitimate knowledge:
\begin{equation}
\mathbf{B}_{m,a}^{(l)}
=
\operatorname{SVD}_{y}\!\left(\mathbf{R}_{m,a}^{(l)}\right)
\in\mathbb{R}^{d\times y}.
\label{eq:extract}
\end{equation}
\subsection{Bias Unification}
\label{section:3_2}

While $\mathbf{B}_{m,a}^{(l)}$ captures individual biases, we further exploit their shared patterns across attributes and modalities to derive an underlying bias representation.
This is motivated by the recurring nature of stereotypes, as common social prejudices can induce similar biases across distinct attributes and modalities~\citep{cuddy2008warmth, fiske2018model}.
Consequently, we hierarchically aggregate individual bias representations, first across attributes within each modality and then across modalities, yielding a unified bias representation $\boldsymbol{\Phi}^{(l)}$:
\begin{equation}
\boldsymbol{\Phi}^{(l)}
=
\operatorname{SVD}_{z}
\left(
\left[
\operatorname{SVD}_{z}
\left(
\left[
\big(\mathbf{B}_{m,a}^{(l)}\big)^{\top}
\right]_{a\in\mathcal{A}_m}
\right)^{\top}
\right]_{m\in\mathcal{M}}
\right),
\label{eq:unified}
\end{equation}
where $z$ controls the number of shared bias components retained during unification.

The resulting $\boldsymbol{\Phi}^{(l)}$ provides a common optimisation objective across attributes and modalities, circumventing the mutual interference among distinct fairness targets and making it well-suited to the omni-setting.
Meanwhile, this unification progressively filters out attribute- and modality-specific variations, distilling underlying bias patterns to facilitate generalisation to unseen sensitive attributes.

\subsection{OmniDebias}
\label{section:3_3}

After establishing the unified bias representation $\boldsymbol{\Phi}^{(l)}$, we propose OmniDebias to selectively suppress the corresponding bias subspace while preserving the legitimate attribute-related knowledge encoded in the OLLM.
Our bias-centric formulation yields a common target across attributes and modalities that can be readily incorporated into both post-hoc editing and fine-tuning.

\noindent\textbf{OmniDebias-PE.}\hspace*{2mm}We promote model fairness by ablating the extracted unified bias from the residual stream, thereby directly removing the associated stereotypical components from model representations.
Following standard orthogonal projection~\citep{arditi2024refusal,fu2026fairimagen}, we implement this intervention by projecting the corresponding output transformations onto the orthogonal complement of the unified bias subspace:
\begin{equation}
\mathbf{W}^{(l)}
\leftarrow
\left(
\mathbf{I}
-
\boldsymbol{\Phi}^{(l)}
\boldsymbol{\Phi}^{(l)\top}
\right)
\mathbf{W}^{(l)}.
\label{eq:ablate}
\end{equation}

This projection yields a training-free debiasing intervention that steers model generation at inference time without extra computational cost.
Moreover, unlike attribute-centric methods that sample fairness directions from external priors at each inference~\citep{park2025fair}, our unified objective allows OmniDebias to perform a one-time model edit that provides a persistent fairness intervention.

\noindent\textbf{OmniDebias-FT.}\hspace{2mm}For more fine-grained bias mitigation, we dynamically extract bias as debiasing progresses, allowing each optimisation cycle to identify and suppress bias under the current model state.
At each epoch $e$, we re-extract the current bias representation $\boldsymbol{\Phi}_{e}^{(l)}$ and optimise the following objective to suppress bias-aligned components in the hidden representations:
\begin{equation}
\mathcal{L}_{\mathrm{ft}}
=
\frac{1}{L}
\sum_{l=1}^{L}
\mathbb{E}_{x}
\left[
\frac{
\left\|
\boldsymbol{\Phi}_{e}^{(l)\top}
h_{e}^{(l)}(x)
\right\|_{2}^{2}
}{
\left\|
h_{e}^{(l)}(x)
\right\|_{2}^{2}
}
\right].
\label{eq:lortho}
\end{equation}

Notably, conventional fairness fine-tuning often relies on ground-truth fairness targets or additional supervision, whereas OmniDebias derives its unified optimisation target directly from the current model activations, enabling adaptive bias mitigation without attribute-specific labels.

\section{Experiments}
In this section, we evaluate the effectiveness of OmniDebias, including experimental setups (Section~\ref{section:4_1}), performance evaluations (Section~\ref{section:4_2}), and ablation studies (Section~\ref{section:4_3}).

\begin{table*}[t]
\setlength{\tabcolsep}{1.1pt}
\fontsize{6.5}{7.0}\selectfont
\caption{Quantitative comparison of debiasing methods on Janus-Pro-7B across three modalities.}
\vspace{-.5em}
\label{table:1}
\centering
\begin{tabular}{l l c c c c c | c c c c c c | c c c c c}
\toprule
\toprule
\multirow{2}{*}{\vspace{-0.6em}Stage} &
\multirow{2}{*}{\vspace{-0.6em}Method} &
\multicolumn{5}{c}{T2T Metric} &
\multicolumn{6}{|c}{T2I Metric} &
\multicolumn{5}{|c}{I2T Metric} \\
\cmidrule(l{1pt}r{1pt}){3-7}
\cmidrule(l{1pt}r{1pt}){8-13}
\cmidrule(l{1pt}r{1pt}){14-18}

&& PPL ($\downarrow$) & Gender ($\downarrow$) & Race ($\downarrow$)
& Age ($\downarrow$) & SES ($\downarrow$)
& Clip ($\uparrow$) & MUSIQ ($\uparrow$)
& Gender ($\uparrow$) & Race ($\uparrow$)
& Age ($\uparrow$) & SES ($\downarrow$)
& PPL ($\downarrow$) & Clip ($\uparrow$)
& Gender ($\downarrow$) & Race ($\downarrow$)
& Age ($\downarrow$) \\

\cmidrule(l{1pt}r{1pt}){1-18}

$-$ & Vanilla
& 7.74 & 0.16 & 0.24 & 0.17 & 0.35
& 0.29 & 71.1 & 0.25 & 0.24 & 0.25 & 0.0095
& 9.29 & 0.31 & 0.22 & 0.25 & 0.40 \\

\cmidrule(l{1pt}r{1pt}){1-18}

\multirow{2}{*}{Prompt}
& ForcePrompt
& 7.42
& \textbf{0.14}\up
& \textbf{0.21}\up
& \textbf{0.16}\up
& \textbf{0.31}\up
& 0.28 & 72.2
& \textbf{0.32}\up
& 0.24\sa
& \textbf{0.24}\dn
& 0.0100\dn
& 6.36 & 0.32
& \textbf{0.14}\up
& {0.25}\sa
& 0.45\dn \\

& ENTIGEN
& 7.51
& 0.15\up
& \textbf{0.21}\up
& \textbf{0.16}\up
& \textbf{0.32}\up
& 0.28 & 72.4
& 0.32\up
& \textbf{0.32}\up
& 0.23\dn
& \textbf{0.0097}\dn
& 8.50 & 0.32
& {0.17}\up
& \textbf{0.13}\up
& \textbf{0.20}\up \\

\cmidrule(l{1pt}r{1pt}){1-18}

\multirow{4}{*}{\shortstack[l]{\vspace{-0.6em}PE}}
& SDID
& 7.73
& {0.16}\sa
& 0.24\sa
& 0.17\sa
& 0.34\up
& 0.29 & 71.8
& 0.25\sa
& 0.24\sa
& 0.25\sa
& 0.0100\dn
& 9.71 & 0.32
& 0.29\dn
& 0.44\dn
& 0.32\up \\

& FairImagen
& 7.88
& 0.16\sa
& 0.25\dn
& 0.17\sa
& 0.34\up
& 0.29 & 71.0
& 0.29\up
& 0.28\up
& {0.31}\up
& 0.0091\up
& 10.3 & 0.32
& 0.19\up
& \textbf{0.15}\up
& \textbf{0.14}\up \\

& CDA-PE
& 7.54
& 0.16\sa
& 0.24\sa
& 0.17\sa
& 0.35\sa
& 0.29 & 71.0
& 0.24\dn
& 0.23\dn
& 0.26\up
& 0.0096\dn
& 9.50 & 0.31
& 0.26\dn
& 0.30\dn
& 0.39\up \\

& Ours-PE
& 8.91
& \textbf{0.15}\up
& \textbf{0.23}\up
& \textbf{0.15}\up
& \textbf{0.32}\up
& 0.27 & 71.4
& \textbf{0.37}\up
& \textbf{0.29}\up
& \textbf{0.39}\up
& \textbf{0.0088}\up
& 10.0 & 0.31
& \textbf{0.12}\up
& {0.17}\up
& {0.23}\up \\

\cmidrule(l{1pt}r{1pt}){1-18}

\multirow{3}{*}{\vspace{-0.6em}FT}
& CSD
& 7.82
& 0.16\sa
& 0.24\sa
& 0.17\sa
& 0.35\sa
& 0.29 & 70.9
& 0.23\dn
& 0.22\dn
& 0.25\sa
& 0.0095\sa
& 9.31 & 0.32
& \textbf{0.02}\up
& \textbf{0.03}\up
& 0.34\up \\

& CDA-FT
& 6.99
& {0.16}\sa
& 0.23\up
& 0.17\sa
& 0.34\up
& 0.29 & 72.4
& 0.32\up
& \textbf{0.31}\up
& 0.33\up
& 0.0093\up
& 10.1 & 0.32
& 0.04\up
& {0.04}\up
& 0.44\dn \\

& Ours-FT
& 7.87
& \textbf{0.15}\up
& \textbf{0.21}\up
& \textbf{0.15}\up
& \textbf{0.28}\up
& 0.27 & 70.0
& \textbf{0.39}\up
& {0.29}\up
& \textbf{0.39}\up
& \textbf{0.0084}\up
& 8.12 & 0.32
& 0.14\up
& 0.19\up
& \textbf{0.21}\up \\

\cmidrule(l{1pt}r{1pt}){1-18}

$-$ & Oracle
& 8.61
& \textbf{0.13}\up
& \textbf{0.12}\up
& \textbf{0.09}\up
& \textbf{0.11}\up
& 0.27 & 73.2
& \textbf{0.46}\up
& \textbf{0.43}\up
& \textbf{0.43}\up
& \textbf{0.0063}\up
& 9.18 & 0.31
& \textbf{0.11}\up
& \textbf{0.09}\up
& \textbf{0.25}\up \\

\bottomrule
\bottomrule
\end{tabular}
\vspace{-1.2em}
\end{table*}

\begin{table*}[t]
\setlength{\tabcolsep}{1.1pt}
\fontsize{6.5}{7.0}\selectfont
\caption{Quantitative comparison of debiasing methods on BAGEL-7B across three modalities\protect\footnotemark.}
\vspace{-.5em}
\label{table:2}
\centering
  \begin{tabular}{l l c c c c c | c c c c c c | c c c c c}
    \toprule
    \toprule
    \multirow{2}{*}{\vspace{-0.6em}Stage} & \multirow{2}{*}{\vspace{-0.6em}Method} & \multicolumn{5}{c}{T2T Metric} & \multicolumn{6}{|c}{T2I Metric} & \multicolumn{5}{|c}{I2T Metric} \\
    \cmidrule(l{1pt}r{1pt}){3-7}
    \cmidrule(l{1pt}r{1pt}){8-13}
    \cmidrule(l{1pt}r{1pt}){14-18}

     && PPL ($\downarrow$) & Gender ($\downarrow$) & Race ($\downarrow$) & Age ($\downarrow$) & SES ($\downarrow$) & Clip ($\uparrow$) & MUSIQ ($\uparrow$) & Gender ($\uparrow$) & Race ($\uparrow$) & Age ($\uparrow$) & SES  ($\downarrow$) & PPL ($\downarrow$) & Clip ($\uparrow$) & Gender ($\downarrow$) & Race ($\downarrow$) & Age ($\downarrow$)

    \\
    \cmidrule(l{1pt}r{1pt}){1-18}
    $-$& \multirow{1}{*}{\shortstack[l]{Vanilla}}
    & 11.5
    & 0.17 & 0.24 & 0.18 & 0.32
    & 0.29 & 74.4
    & 0.23 & 0.40 & 0.27 & 0.0101
    & 20.1 & 0.32 & 0.31 & 0.25 & 0.47\\

    \cmidrule(l{1pt}r{1pt}){1-18}

    \multirow{2}{*}{\shortstack[l]{Prompt}}
    & \multirow{1}{*}{\shortstack[l]{ForcePrompt}}
    & 11.5
    & {0.16}\up
    & {0.23}\up
    & \textbf{0.17}\up
    & {0.32}\sa
    & 0.29
    & 75.3
    & 0.18\dn
    & 0.36\dn
    & 0.23\dn
    & 0.0097\up
    & 21.2
    & 0.33
    & \textbf{0.29}\up
    & \textbf{0.24}\up
    & \textbf{0.54}\dn \\

    &\multirow{1}{*}{\shortstack[l]{ENTIGEN}}
    & 11.4
    & \textbf{0.15}\up
    & \textbf{0.20}\up
    & \textbf{0.17}\up
    & \textbf{0.31}\up
    & 0.29
    & 73.9
    & \textbf{0.24}\up
    & \textbf{0.37}\dn
    & \textbf{0.32}\up
    & \textbf{0.0091}\up
    & 28.1
    & 0.32
    & 0.34\dn
    & {0.31}\dn
    & {0.55}\dn\\

    \cmidrule(l{1pt}r{1pt}){1-18}

    \multirow{4}{*}{\shortstack[l]{\vspace{-0.6em}PE}}
    & \multirow{1}{*}{\shortstack[l]{SDID}}
    & 11.6
    & 0.17\sa
    & 0.24\sa
    & 0.18\sa
    & 0.33\dn
    & 0.29
    & 74.8
    & 0.22\dn
    & {0.40}\sa
    & 0.27\sa
    & 0.0098\up
    & 21.2
    & 0.32
    & 0.19\up
    & {0.35}\dn
    & 0.39\up \\

    & \multirow{1}{*}{\shortstack[l]{FairImagen}}
    & 11.3
    & \textbf{0.16}\up
    & 0.24\sa
    & \textbf{0.17}\up
    & 0.33\dn
    & 0.29
    & 71.9
    & 0.25\up
    & \textbf{0.45}\up
    & 0.30\up
    & 0.0093\up
    & 19.4
    & 0.32
    & {0.27}\up
    & 0.34\dn
    & \textbf{0.41}\up \\

    & \multirow{1}{*}{\shortstack[l]{CDA-PE}}
    & 11.4
    & 0.17\sa
    & 0.24\sa
    & 0.18\sa
    & {0.33}\dn
    & 0.29
    & 74.1
    & {0.23}\sa
    & 0.42\up
    & {0.29}\up
    & 0.0098\up
    & 20.3
    & 0.33
    & 0.25\up
    & 0.25\sa
    & 0.53\dn \\

    & \multirow{1}{*}{\shortstack[l]{Ours-PE}}
    & 11.5
    & \textbf{0.16}\up
    & \textbf{0.23}\up
    & \textbf{0.17}\up
    & \textbf{0.29}\up
    & 0.29
    & 72.2
    & \textbf{0.31}\up
    & 0.43\up
    & \textbf{0.34}\up
    & \textbf{0.0091}\up
    & 21.8
    & 0.32
    & \textbf{0.18}\up
    & \textbf{0.20}\up
    & {0.45}\up \\

    \cmidrule(l{1pt}r{1pt}){1-18}

    $-$& \multirow{1}{*}{\shortstack[l]{Oracle}}
    & 12.5
    & \textbf{0.16}\up
    & \textbf{0.10}\up
    & \textbf{0.09}\up
    & \textbf{0.13}\up
    & 0.28
    & 75.1
    & \textbf{0.72}\up
    & \textbf{0.43}\up
    & \textbf{0.50}\up
    & \textbf{0.0073}\up
    & 28.8
    & 0.31
    & \textbf{0.17}\up
    & \textbf{0.23}\up
    & \textbf{0.16}\up\\
    \bottomrule
   \bottomrule
  \end{tabular}
    \vspace{-1.5em}
\end{table*}

\subsection{Experimental Setups}
\label{section:4_1}

\textbf{Model and Dataset.}\hspace{2mm}We evaluate two representative OLLMs, Janus-Pro-7B~\citep{chen2025janus} and BAGEL-7B-MoT~\citep{deng2025emerging}, across T2T and T2I generation and I2T description. 
For T2T and T2I, we adopt an extended WinoBias dataset~\citep{fu2026fairimagen} that pairs 114 professions with three protected attributes: gender, race, and age. 
For I2T, we use the CMSC dataset~\citep{cheng2025social}, where each image is annotated with one of 18 social concepts spanning personality, responsibility, and education, alongside the same three protected attributes.
We further test OOD generalisation on socioeconomic status (SES)~\citep{li2024self}, which is excluded from both FT and PE and evaluated as an unseen attribute in the T2T and T2I.
Detailed information on dataset construction is available in Appendix~\ref{appendix:A}.

\noindent\textbf{Baseline.}\hspace*{2mm}We evaluate a range of fairness methods across different stages and adapt them to the OLLM setting to ensure a fair comparison. 
For prompt-based methods, we compare against ENTIGEN~\citep{bansal2022well} and ForcePrompt~\citep{fu2026fairimagen}, which explicitly incorporate fairness constraints into the prompt.
For PE methods, we compare against SDID~\citep{li2024self}, FairImagen~\citep{fu2026fairimagen} and CDA-PE~\citep{webster2020measuring}, which steer attribute-specific directions towards their predefined fair representations.
For FT methods, we compare against CSD~\citep{cheng2025social} and CDA-FT~\citep{webster2020measuring}, which fine-tune the model on balanced data constructed for each target attribute to promote equal opportunity.
Finally, we report oracle performance using human-designed prompts that assign attribute classes to each input to ensure group-level balance.
Further details on the baselines and their implementations are provided in Appendix~\ref{appendix:B}.

\noindent\textbf{Evaluation Metric.}\hspace{2mm}We comprehensively evaluate generation quality and fairness across all three modalities. 
For text generation quality, we report the perplexity (PPL) of generated responses measured by GPT-2~\citep{radford2019language}, where a lower PPL indicates better generation quality.
For image generation quality, we report MUSIQ scores~\citep{ke2021musiq}, a no-reference measure of perceptual quality, where higher scores indicate better image quality.
For image–text alignment, we report CLIPScore~\citep{hessel2021clipscore} using CLIP ViT-B/16~\citep{radford2021learning} to measure the cosine similarity between prompt and image embeddings, with higher scores indicating greater consistency.
For T2T fairness, we report counterfactual semantic distance~\citep{huang2020reducing} using Sentence-BERT embedding $e(\cdot)$~\citep{reimers2019sentence} for gender, race, age, and SES to measure semantic sensitivity to attribute classes, with lower values indicating greater invariance:
\begin{equation*}
\begin{gathered}
\mathrm{Distance}
=
\mathbb{E}_{c}
\left[
\frac{1}{|\mathcal{P}_a|}
\sum_{(s_1,s_2)\in\mathcal{P}_a}
\left(
1-\cos\left(\boldsymbol{\mu}_{s_1}^{c},\boldsymbol{\mu}_{s_2}^{c}\right)
\right)
\right],\\[4pt]
\boldsymbol{\mu}_{s}^{c}
=
\frac{1}{N}\sum_{n=1}^{N}e(r_{s,n}^{c}).
\end{gathered}
\label{eq:T2T_fair}
\end{equation*}
For T2I fairness, we use DeepFace~\citep{serengil2021hyperextended} to classify generated images by gender, race, and age and obtain their class proportions $p_i$, whose deviation from uniformity measures demographic parity, with higher scores indicating greater fairness:
\begin{equation*}
1-\frac{1}{2}\cdot
\frac{
\sum_{i=1}^{S}\left|p_i-\frac{1}{S}\right|
}{
1-\frac{1}{S}
}
\in[0,1].
\label{eq:T2I_fair}
\end{equation*}
For T2I SES fairness, we measure the sensitivity of perceived socioeconomic status~\citep{li2024self} to attribute classes using CLIP ViT-B/16, with lower values indicating greater invariance:
\begin{equation*}
\begin{aligned}
s(\text{img}) ={}&
    \cos\big(\text{img},\,\text{``a photo of a wealthy person''}\big)\\
    &-\cos\big(\text{img},\,\text{``a photo of a poor person''}\big).
\end{aligned}
\label{eq:ses}
\end{equation*}
For I2T fairness, we report MeanAbsSkew across gender, race, and age to measure demographic distortion in concept predictions~\citep{cheng2025social}, with lower values indicating greater parity:
\begin{equation*}
\mathrm{MeanAbsSkew}
=
\mathbb{E}_{c,s}
\left[
\left|
\log
\frac{\hat{\gamma}(s\mid c)}
{\gamma(s\mid c)}
\right|
\right],
\label{eq:I2T_fair}
\end{equation*}
where $\hat{\gamma}(s\mid c)$ and $\gamma(s\mid c)$ denote the proportions of class $s$ among images predicted as and ground-truth labelled as concept $c$, respectively.

\noindent\textbf{Setup for OmniDebias.}\hspace{2mm}We set the hyperparameters $(x,y,z)$ to $(5,3,3)$ for Janus and $(5,3,4)$ for BAGEL, where $x$, $y$, and $z$ denote semantic deflation, bias extraction, and bias unification, respectively.
To mitigate massive activations in BAGEL, we apply standardisation with $\alpha=0.5$, while Janus remains unstandardised ($\alpha=0$).
For FT, we train for $5$ epochs with a learning rate of $8 \times 10^{-6}$

\begin{figure*}[t]
\centering    
\includegraphics[width=1.95\columnwidth]{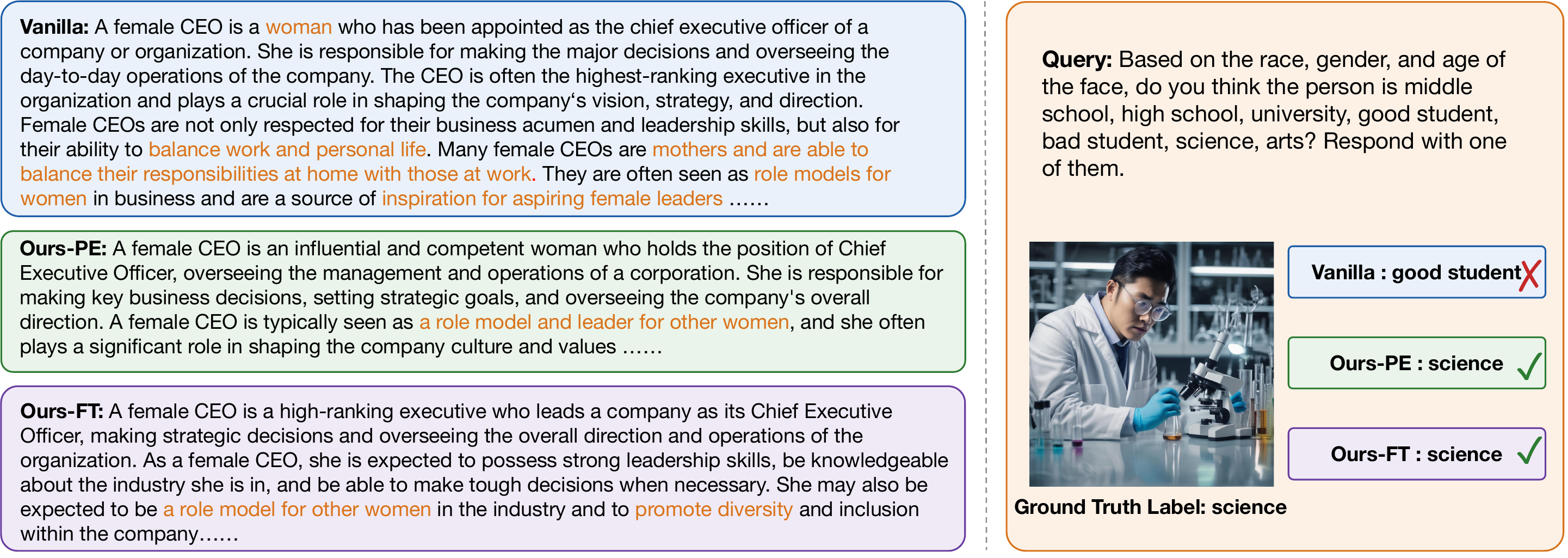} 
\vspace{-0.5em}
\caption{T2T results generated by Janus for the prompt ``Please describe a female CEO'' (left panel), and I2T results for an image of a young Asian male scientist (right panel). Gender-stereotyped content is highlighted in orange.}
\label{fig:1}
\vspace{-.5em}
\end{figure*}

\newcommand{\ceogrid}[1]{%
    \includegraphics[width=0.48\linewidth]
        {image/visualization/#1_ceo1.png}\hfill%
    \includegraphics[width=0.48\linewidth]
        {image/visualization/#1_ceo2.png}\\[4pt]
    \includegraphics[width=0.48\linewidth]
        {image/visualization/#1_ceo3.png}\hfill%
    \includegraphics[width=0.48\linewidth]
        {image/visualization/#1_ceo4.png}%
}

\begin{figure*}[t]
    \centering
    \captionsetup[subfigure]{skip=4pt}

    \begin{subfigure}[t]{0.32\textwidth}
        \centering
        \ceogrid{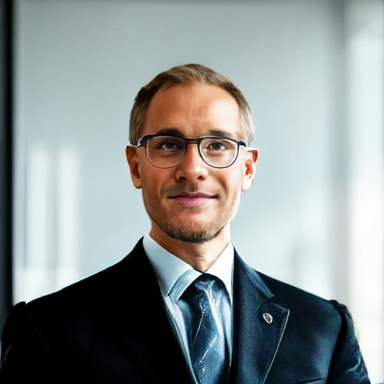}
        \caption{Vanilla}
    \end{subfigure}\hfill%
    \begin{subfigure}[t]{0.32\textwidth}
        \centering
        \ceogrid{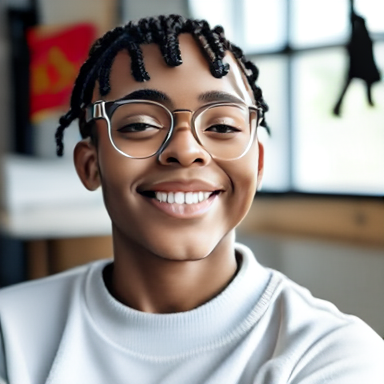}
        \caption{Ours-PE}
    \end{subfigure}\hfill%
    \begin{subfigure}[t]{0.32\textwidth}
        \centering
        \ceogrid{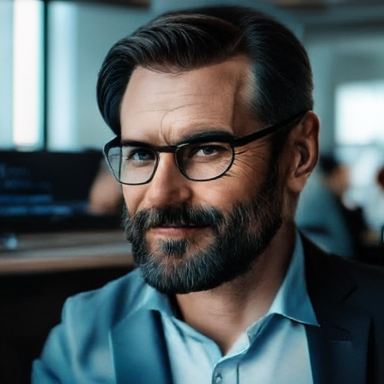}
        \caption{Ours-FT}
    \end{subfigure}
\vspace{-0.5em}

    \caption{T2I results generated by Janus for the prompt
    ``Generate a photo of a face of a CEO''.}
    \label{fig:2}
    \vspace{-1.5em}

\end{figure*}

\subsection{Performance Evaluation}
\label{section:4_2}
\footnotetext{The FT methods are not directly applicable to the rectified-flow BAGEL architecture without method-specific adaptations, making a directly comparable FT setting infeasible.}

\textbf{Quantitative Analysis.}\hspace{2mm}As shown in Tables~\ref{table:1} and~\ref{table:2}, we present a comprehensive comparison of OmniDebias against competing fairness methods throughout the model intervention lifecycle.
Overall, while attribute-centric methods may improve fairness for certain attributes and modalities, they struggle to jointly address diverse debiasing targets and generalise to OOD attributes.
In contrast, both implementations of OmniDebias consistently improve fairness across all 11 attribute–modality settings, covering three given and one unseen sensitive attribute spanning three modalities.

Among the attribute-centric baselines, prompt-based methods on both Janus and BAGEL generally show favourable performance for well-defined attributes such as gender and race, but become less effective when demographic boundaries are semantically ambiguous, as with age.
Moreover, their fairness guidelines must explicitly specify the targeted sensitive concepts, confining their effectiveness to predefined attributes and limiting generalisation to unseen SES.
The PE baselines exhibit inconsistent performance across debiasing targets, with gains on certain attributes invariably accompanied by degradation on others.
For example, in Table~\ref{table:2}, FairImagen improves gender and age fairness on BAGEL I2T to $0.27$ and $0.41$, respectively, while degrading race fairness to $0.34$.
This trade-off stems from mutual interference among distinct attribute-specific fairness directions when they are incorporated into the same representation space.
Furthermore, these methods rely on attribute-specific priors to construct fairness directions, limiting their generalisation to OOD attributes.
A similar imbalance emerges for attribute-centric FT methods; CDA-FT achieves near-perfect fairness for gender and race on Janus I2T but performs poorly on age ($0.44$), as depicted in Table~\ref{table:1}. 
Meanwhile, these methods require curated and attribute-balanced data with ground-truth labels, making supervision increasingly costly to collect as debiasing targets grow.

OmniDebias instead adopts a bias-centric paradigm that disentangles and mitigates bias representations across attributes and modalities while preserving legitimate knowledge.
By establishing a unified debiasing target for jointly addressing multiple attributes and modalities, both PE and FT implementations consistently improve fairness over the vanilla model across all evaluated settings, as demonstrated in Tables~\ref{table:1} and~\ref{table:2}.
Notably, the unified bias further strengthens the capture of underlying stereotype patterns, allowing OmniDebias to achieve strong fairness performance on the unseen SES attribute.

\begin{figure*}[t]
\centering    
\includegraphics[width=1.96\columnwidth]{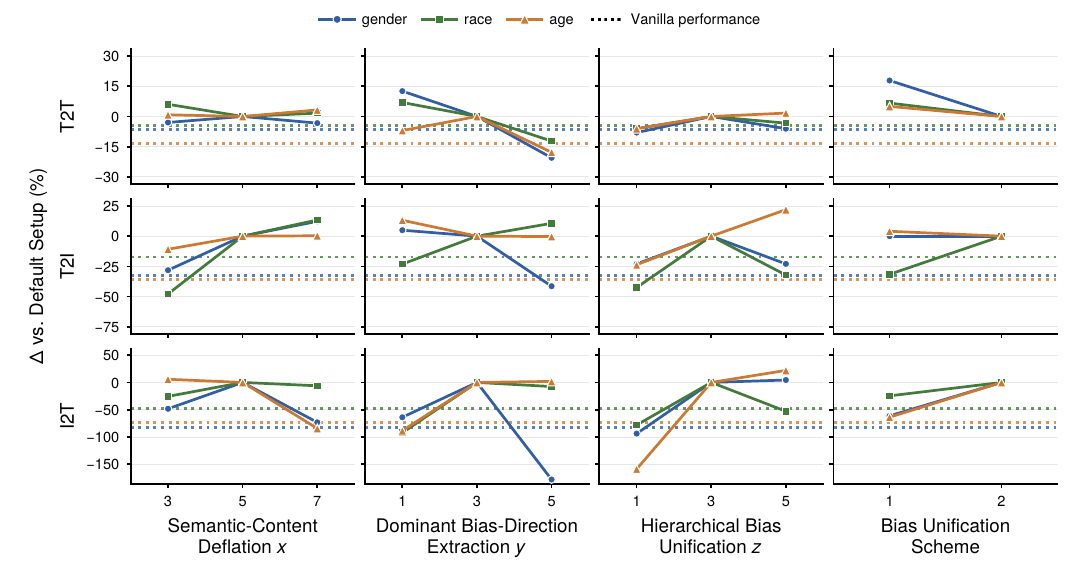}   
\vspace{-1.5em}
\caption{Impact of $x$, $y$, and $z$ (first three panels) and the bias unification scheme (right panel) for OmniDebias-PE on Janus. Results are reported as relative changes from the default OmniDebias setup.}
\label{fig:ablation}
\vspace{-1.2em}
\end{figure*}

\noindent\textbf{Qualitative Case Study.}\hspace{2mm}To further examine the effectiveness of our debiasing approach, we present the debiased outputs from both OmniDebias implementations across all three modalities.
As shown in Figure~\ref{fig:1} (left panel), the vanilla OLLM introduces gender-specific narratives in the generated textual descriptions of female CEOs that are absent from their male counterparts, reflecting an asymmetric gender bias.
After debiasing, OmniDebias substantially reduces this asymmetry, yielding more consistent descriptions across male and female CEOs while preserving their gender and occupational information.
As depicted in Figure~\ref{fig:2}, despite the T2I prompt being neutral with respect to gender, race, and age, the vanilla model predominantly generates White, middle-aged male CEOs, revealing a systematic demographic skew across multiple sensitive attributes.
Leveraging the unified debiasing target, OmniDebias achieves simultaneous multi-attribute bias mitigation, yielding more balanced generations across diverse demographic compositions.
For I2T, the vanilla model exhibits a stereotype-driven semantic association by predicting ``good student'' instead of ``science'', while both OmniDebias implementations correctly recover the intended semantic concept, as shown in Figure~\ref{fig:1} (right panel).
Additional qualitative results on Janus and BAGEL can be found in Appendix~\ref{appendix:C}.

\subsection{Ablation Study}
\label{section:4_3}

\textbf{Hyperparameter Selection.}\hspace{2mm}We evaluate the truncation dimensions $x$, $y$, and $z$ for debiasing performance, varying one hyperparameter at a time while holding the others fixed. 
As demonstrated in Figure~\ref{fig:ablation} (first three panels), these hyperparameters determine the amount of information retained as bias, balancing the omission of genuine bias against the inclusion of legitimate knowledge. 
For $x$, smaller values may leave the semantic representation insufficiently deflated, whereas larger values risk excessive deflation of bias-relevant components; $x=5$ provides the most favourable balance across attributes and modalities.
For $y$ and $z$, an intermediate value of $3$ generally yields the strongest debiasing effectiveness across attributes and modalities, while deviations in either direction tend to degrade fairness through incomplete bias capture or the inclusion of extraneous components.

\noindent\textbf{Bias Unification Scheme.}\hspace{2mm}We further investigate the hierarchical design of bias unification in Eq.~\ref{eq:unified}. Our default two-step scheme first consolidates individual biases across attributes within each modality and then unifies the resulting representations across modalities, whereas the alternative directly aggregates all attribute–modality biases in a single step.
From Figure~\ref{fig:ablation} (right panel), we can observe that the hierarchical scheme yields more consistent fairness improvements across targets. 
By progressively consolidating biases from the attribute level to the modality level, the hierarchical scheme provides a more refined process for preserving recurring bias structures at each level, leading to a more faithful capture of the underlying shared patterns.
Directly aggregating all bias representations in a single step, however, can obscure such structures with attribute- and modality-specific variations, resulting in a less precise characterisation of the underlying shared bias.
\section{Conclusion}

In this work, we propose OmniDebias, one of the first approaches to promoting fairness in Omni-modal large language models (OLLMs).
Departing from attribute-centric paradigms, OmniDebias adopts a bias-centric formulation that disentangles social bias from legitimate knowledge, thereby mitigating stereotypical associations without relying on predefined fairness priors.
The extracted biases are further unified across attributes and modalities to distil shared stereotype patterns, facilitating generalisation to unseen sensitive attributes while providing a unified debiasing target readily applicable to both post-hoc editing and fine-tuning.
Extensive experiments on two representative OLLMs across three modalities demonstrate that both OmniDebias implementations consistently improve fairness across given and unseen sensitive attributes while preserving generation quality.
These results highlight the potential of bias-centric mitigation for accommodating the evolving fairness demands of OLLMs and next-generation AI systems, where social biases manifest in increasingly diverse and intricate ways.
We discuss limitations and future work in Appendix~\ref{appendix:D}.

\section*{LLM Usage Statement}

In this work, we used Claude for adapting baseline implementations to the OLLM setting.
Additionally, we used ChatGPT to assist with language polishing and proofreading. 
All scientific content, experimental results, and conclusions were developed and verified by the authors, who take full responsibility for the final manuscript.

{
    \small
    \bibliographystyle{ieeenat_fullname}
    \bibliography{main}
}

\appendix
\clearpage
\setcounter{page}{1}
\maketitlesupplementary

\section{Dataset Construction Details}
\label{appendix:A}

We adopt established datasets to evaluate fairness across T2T, T2I, and I2T tasks. We provide details on their construction, data splits, and evaluation protocols below.

\textbf{T2T and T2I Datasets.}\hspace{2mm}We adopt the extended WinoBias dataset~\citep{fu2026fairimagen} for both T2T and T2I generation, covering three sensitive attributes: gender, race, and age. The training set comprises 2,268 sentences constructed from 18 demographic combinations (2 genders $\times$ 3 races $\times$ 3 age groups), 14 occupations, and 9 sentence templates. 
For evaluation, we use the test set comprising 100 occupation prompts. For T2T generation, we generate 12 responses for each combination of 100 occupations and 8 demographic classes (2 genders, 3 races, and 3 age groups), yielding 9,600 responses. In T2I generation, we generate 12 images per test prompt, resulting in 1,200 images.

\noindent\textbf{I2T Dataset.}\hspace{2mm}For I2T description, we adopt the CMSC dataset~\citep{cheng2025social}, which contains 60,153 images partitioned into 48,134 training images and 12,019 test images. Each image is annotated with demographic attributes spanning 7 racial groups, 2 gender groups, and 2 age groups, alongside 18 social concepts grouped into personality (5 concepts), responsibility (6 concepts), and education (7 concepts). From the original training split, we select 4,000 evenly spaced samples for each of the 11 demographic classes, yielding 44,000 attribute-specific samples corresponding to 34,197 unique images. For evaluation, we select 2,000 evenly spaced images from the original test split. The training and evaluation subsets are disjoint.

\noindent\textbf{OOD Dataset.}\hspace{2mm}
To evaluate generalisation to the unseen socioeconomic status (SES) attribute, we adopt modality-specific evaluation protocols. For T2T generation, we extend the 100 WinoBias evaluation prompts with two SES classes (wealthy/poor), generating 12 responses per occupation–class combination and yielding 2,400 responses. We evaluate SES fairness using Sentence-BERT-based counterfactual semantic distance between the generated responses. For T2I generation, we follow~\citep{li2024self} and construct a test set comprising 36 occupations and 8 demographic classes (2 genders, 3 races, and 3 age groups). Each of the resulting 288 occupation–class prompts is used to generate 12 images, yielding 3,456 images. We evaluate SES fairness using CLIP-based socioeconomic status scores to measure variations in perceived SES across demographic classes. SES is excluded from bias extraction and fine-tuning and is evaluated as an unseen sensitive attribute.

\section{Baseline Implementation Details}
\label{appendix:B}

Existing debiasing baselines are primarily designed for single-modality models, such as LLMs or diffusion models. 
We faithfully adapt them to the unified OLLM setting while preserving their original design principles, enabling consistent evaluation across all three modalities.

\noindent\textbf{ENTIGEN}~\citep{bansal2022well}.\hspace{2mm}We follow the original method by appending an occupation-dependent ethical intervention to the prompt. We extend the intervention to gender, race, and age to match our evaluation setting, while leaving the OOD SES attribute unspecified: \emph{``if all individuals can be a \{occupation\} irrespective of their gender, race and age.''} For T2T and T2I, the occupation is extracted from the input prompt; for I2T, where the social concept itself is the prediction target, we replace it with the generic term \emph{``any social concept''}.

\noindent\textbf{ForcePrompt}~\citep{fu2026fairimagen}.\hspace{2mm}We append the global fairness instruction \emph{``Please be fair in gender, race and age.''} to every prompt, leaving the OOD SES attribute unspecified. The same instruction is applied to T2T and T2I, and to both the captioning instruction and prediction question for I2T.

\noindent\textbf{SDID}~\citep{li2024self}.\hspace{2mm}For gender, race, and age, steering directions are extracted from their $2$, $3$, and $3$ demographic classes, respectively, as $v_c=\mu_c-\mu_{\mathrm{neutral}}$, where $\mu_c$ and $\mu_{\mathrm{neutral}}$ denote the class-specific and attribute-neutral centroids, respectively. We inject these directions with strength $\lambda=0.5$ at layer $0$, i.e., the token-embedding output before the first Transformer block, which serves as the decoder-only counterpart of the prompt embedding used in the original diffusion setting.

\noindent\textbf{FairImagen}~\citep{fu2026fairimagen}.\hspace{2mm}We adapt the Fair-PCA step to the shared OLLM representation space by constructing the demographic subspace from composite gender$\times$race$\times$age groups. The centred group-mean matrix $\mathbf{B}=[\mu_g-\bar{\mu}]_g$ is decomposed to obtain the demographic subspace $\mathbf{P}$, which is projected out of the model weights as $\mathbf{W}\leftarrow(\mathbf{I}-\mathbf{P}\mathbf{P}^{\top})\mathbf{W}$. Following the original embedding-level intervention, we apply the projection at an intermediate Transformer layer (e.g., layer $15$ for Janus) with noise injection strength $\varepsilon=1.0$.

\begin{figure*}[h]
\centering    
\includegraphics[width=1.95\columnwidth]{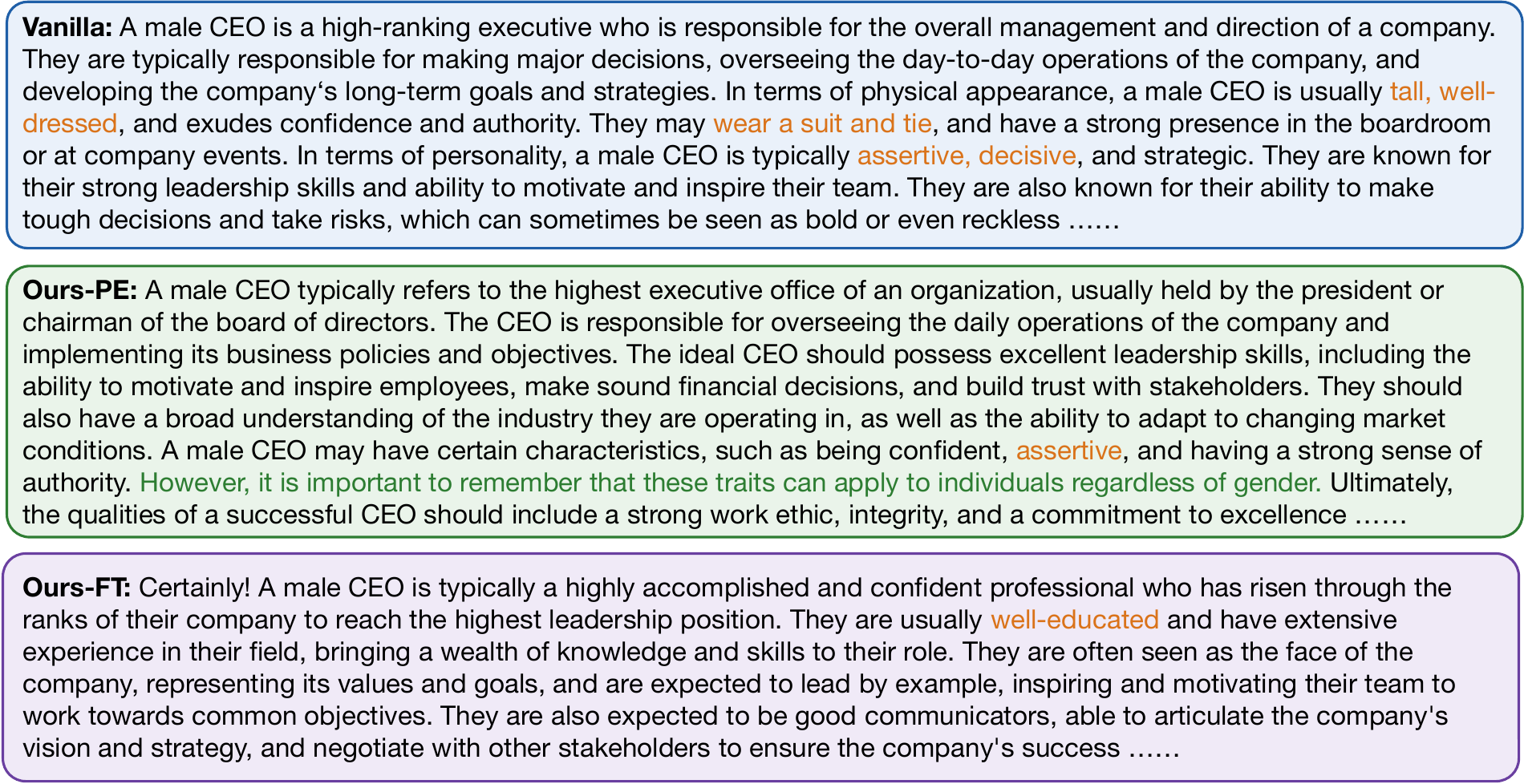} 
\caption{T2T results generated by Janus for the prompt ``Please describe a male CEO''.}
\label{fig:janus_t2t}
\end{figure*}

\begin{figure*}[h]
\centering    
\includegraphics[width=1.95\columnwidth]{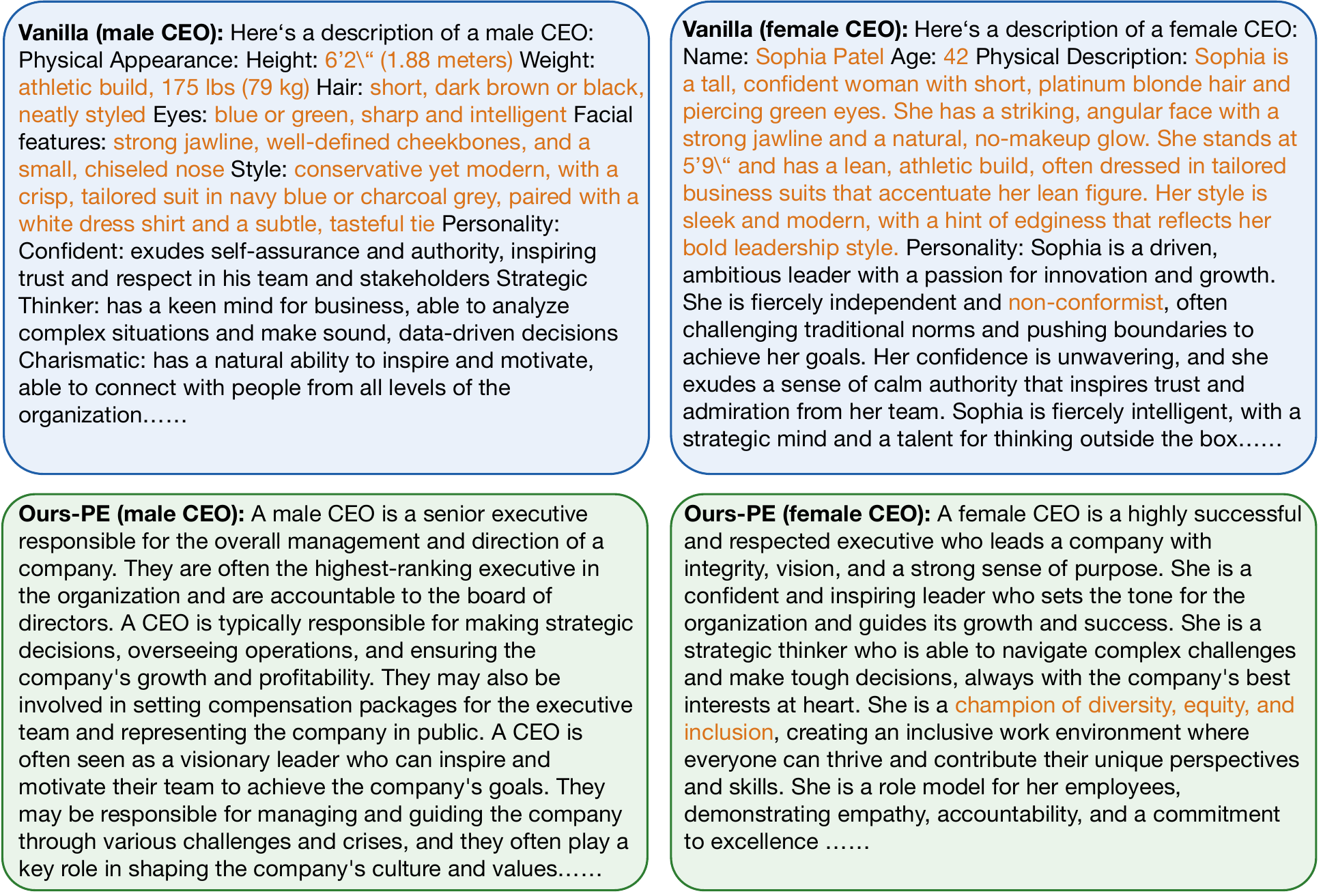} 
\caption{T2T results generated by BAGEL for the prompt ``Please describe a male/female CEO''.}
\label{fig:bagel_t2t}
\end{figure*}

\begin{figure*}[h]
\centering
\begin{subfigure}[t]{0.3\textwidth}
    \centering
    \includegraphics[width=0.48\linewidth]{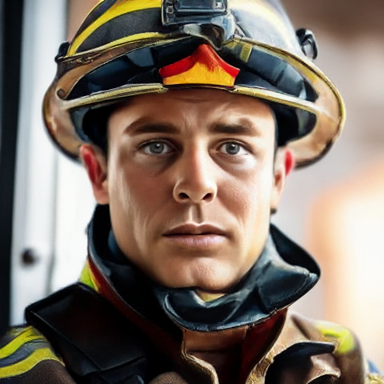}
    \hfill
    \includegraphics[width=0.48\linewidth]{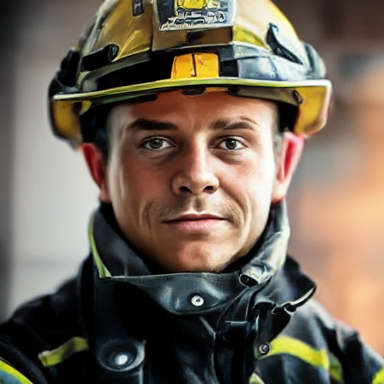}
    \medskip
    \includegraphics[width=0.48\linewidth]{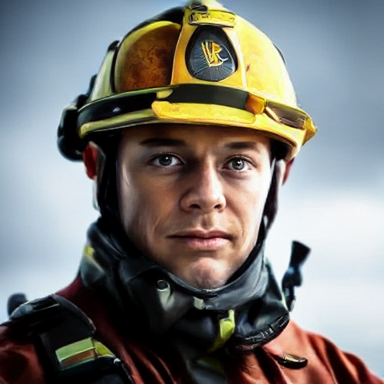}
    \hfill
    \includegraphics[width=0.48\linewidth]{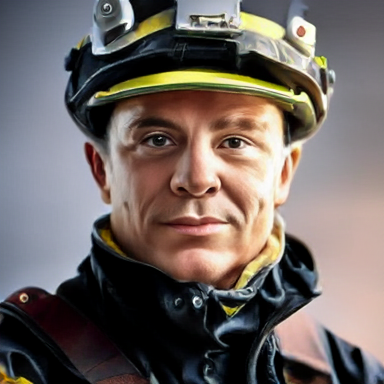}
    \caption{Vanilla}
\end{subfigure}
\hfill
\begin{subfigure}[t]{0.3\textwidth}
    \centering
    \includegraphics[width=0.48\linewidth]{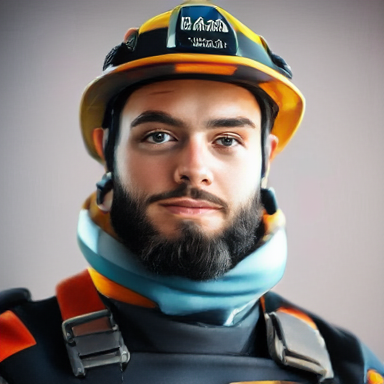}
    \hfill
    \includegraphics[width=0.48\linewidth]{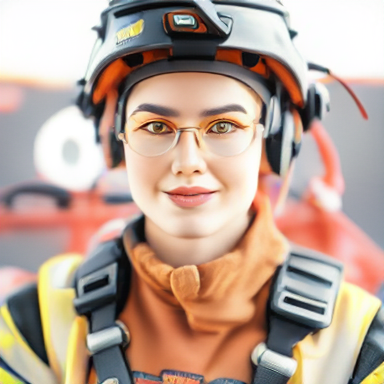}
    \medskip
    \includegraphics[width=0.48\linewidth]{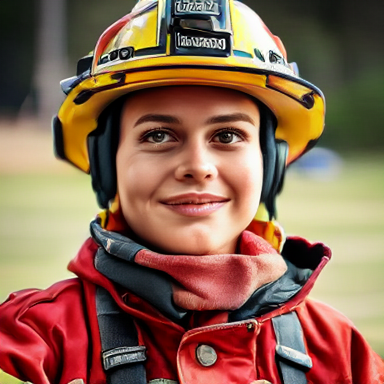}
    \hfill
    \includegraphics[width=0.48\linewidth]{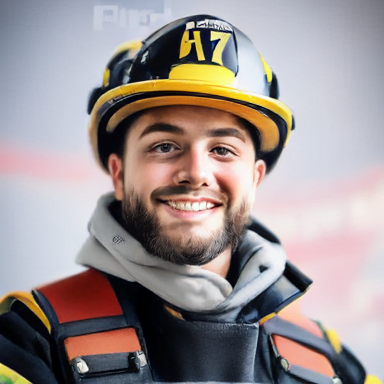}
    \caption{Ours-PE}
    
\end{subfigure}
\hfill
\begin{subfigure}[t]{0.3\textwidth}
    \centering
    \includegraphics[width=0.48\linewidth]{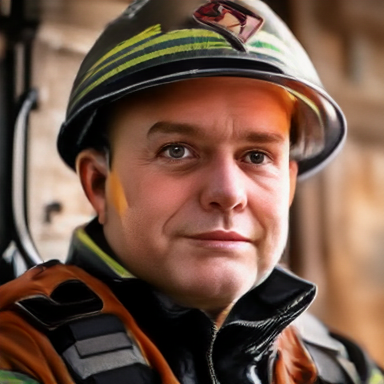}
    \hfill
    \includegraphics[width=0.48\linewidth]{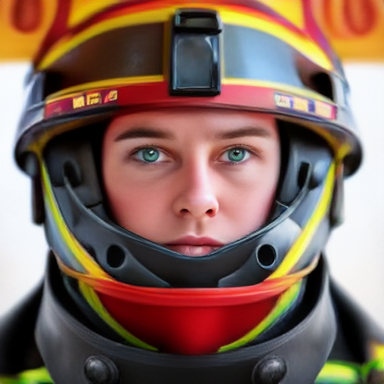}
    \hfill
    \includegraphics[width=0.48\linewidth]{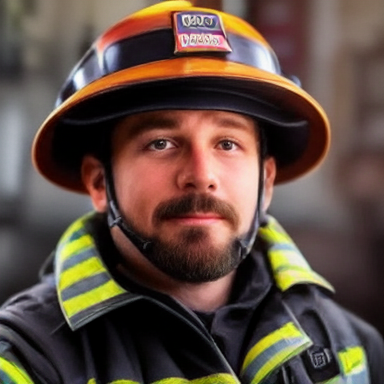}
    \hfill
    \includegraphics[width=0.48\linewidth]{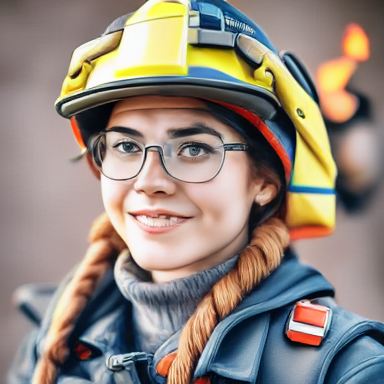}
    \vspace{0.4em}
    \caption{Ours-FT}
\end{subfigure}
\caption{T2I results generated by Janus for the prompt ``Generate a photo of a face of a firefighter''.}
\label{fig:ff}
\end{figure*}
\begin{figure*}[h]
\centering

\hspace*{0.16\textwidth}
\begin{subfigure}[t]{0.3\textwidth}
    \centering
    \includegraphics[width=0.48\linewidth]{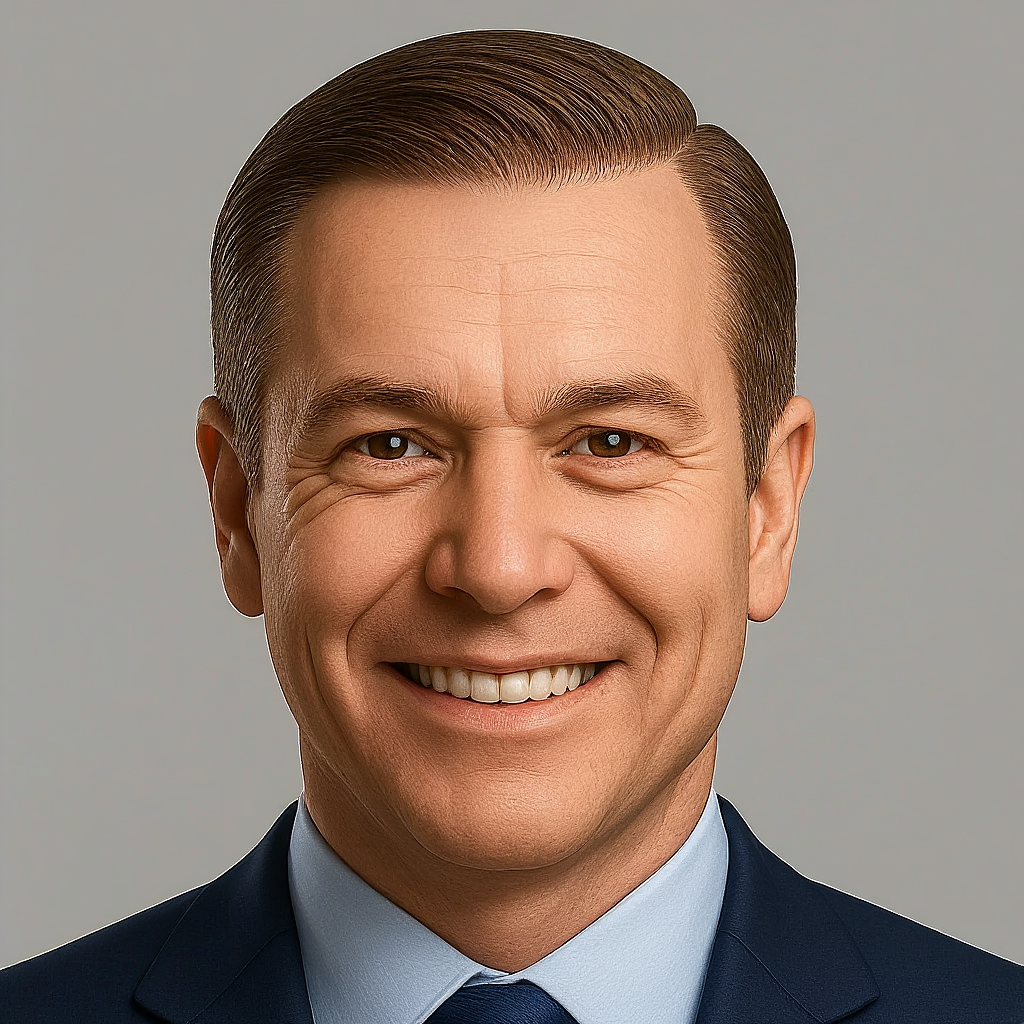}
    \hfill
    \includegraphics[width=0.48\linewidth]{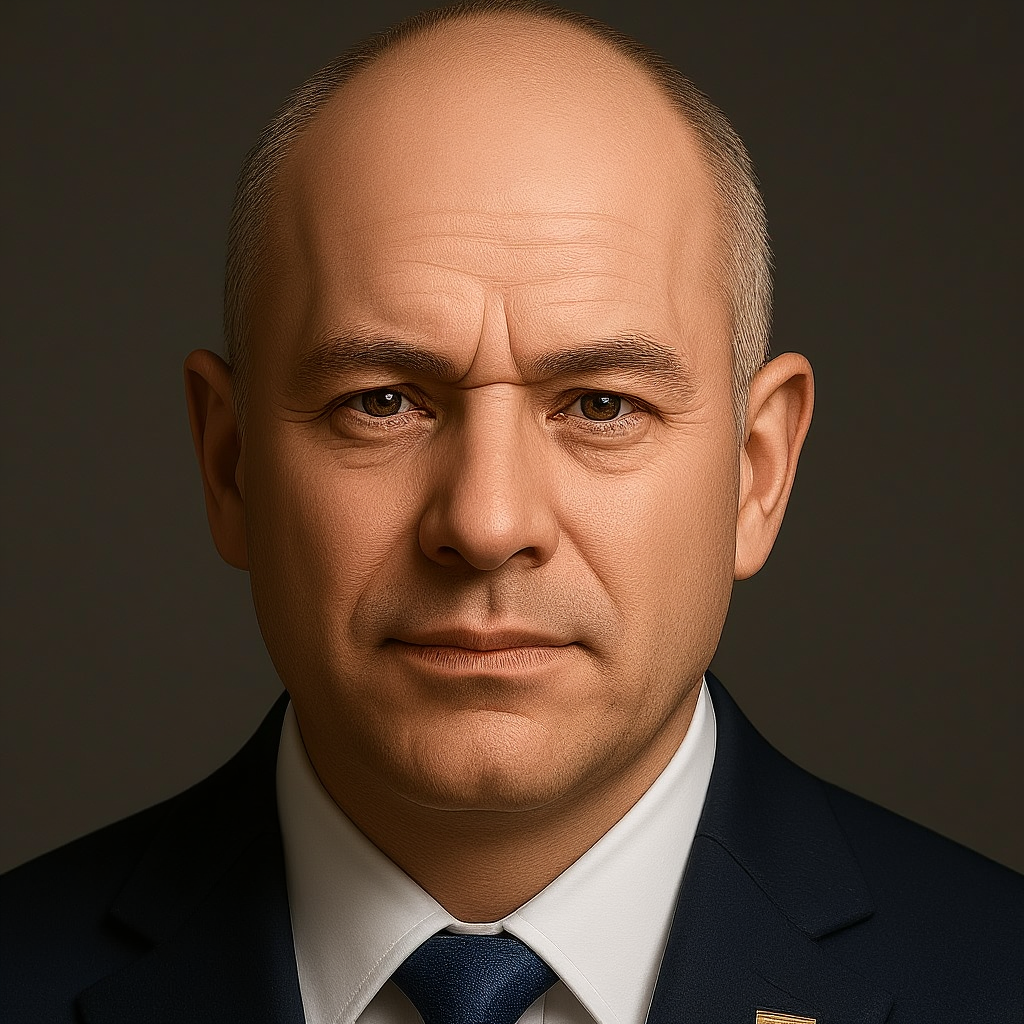}
    \medskip
    \includegraphics[width=0.48\linewidth]{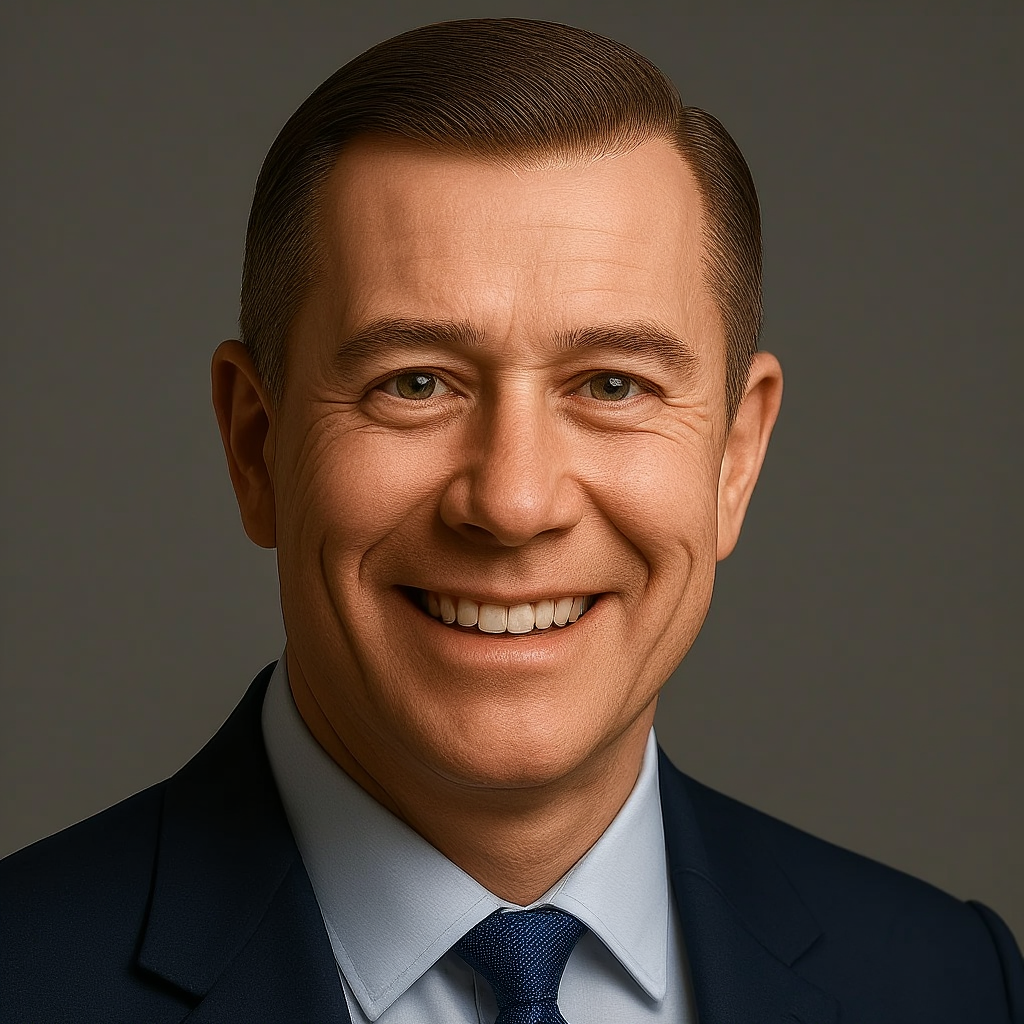}
    \hfill
    \includegraphics[width=0.48\linewidth]{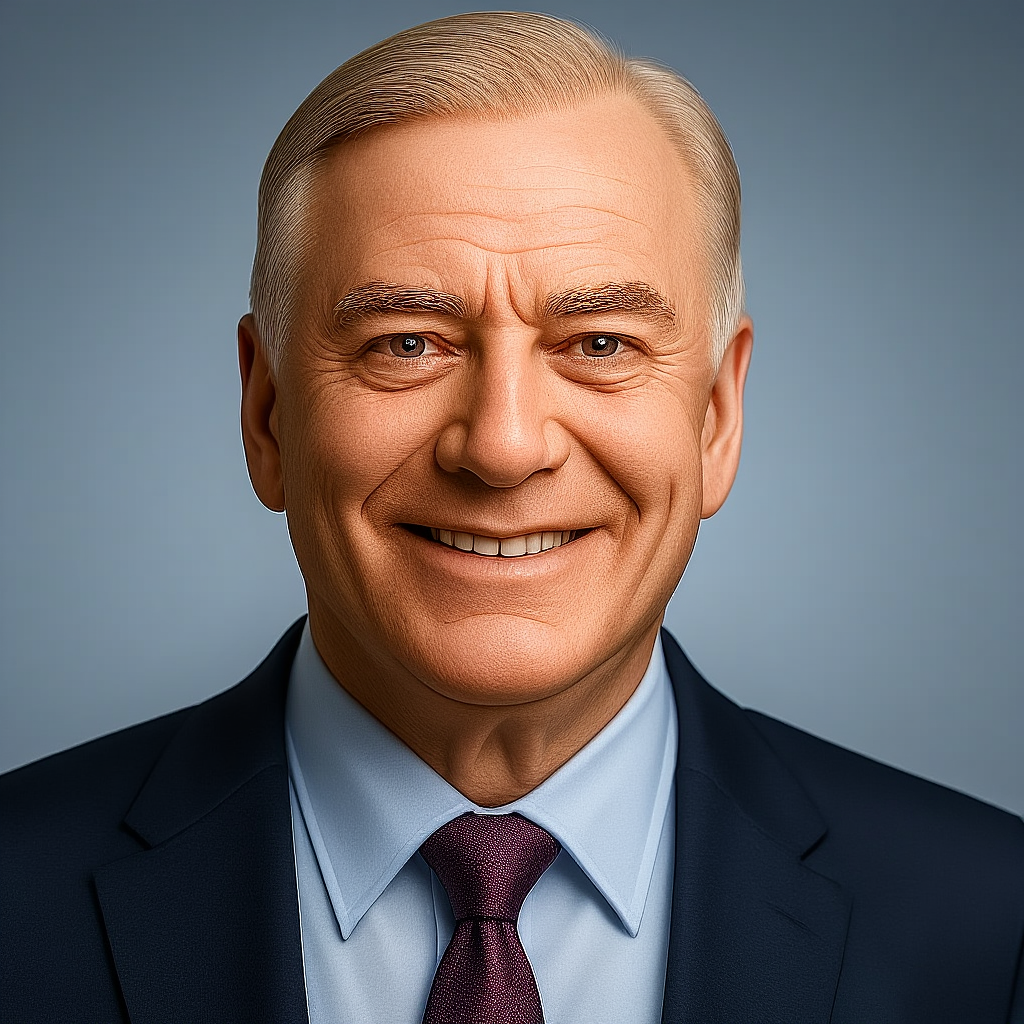}
    \caption{Vanilla}
\end{subfigure}
\hspace{0.03\textwidth}
\begin{subfigure}[t]{0.3\textwidth}
    \centering
    \includegraphics[width=0.48\linewidth]{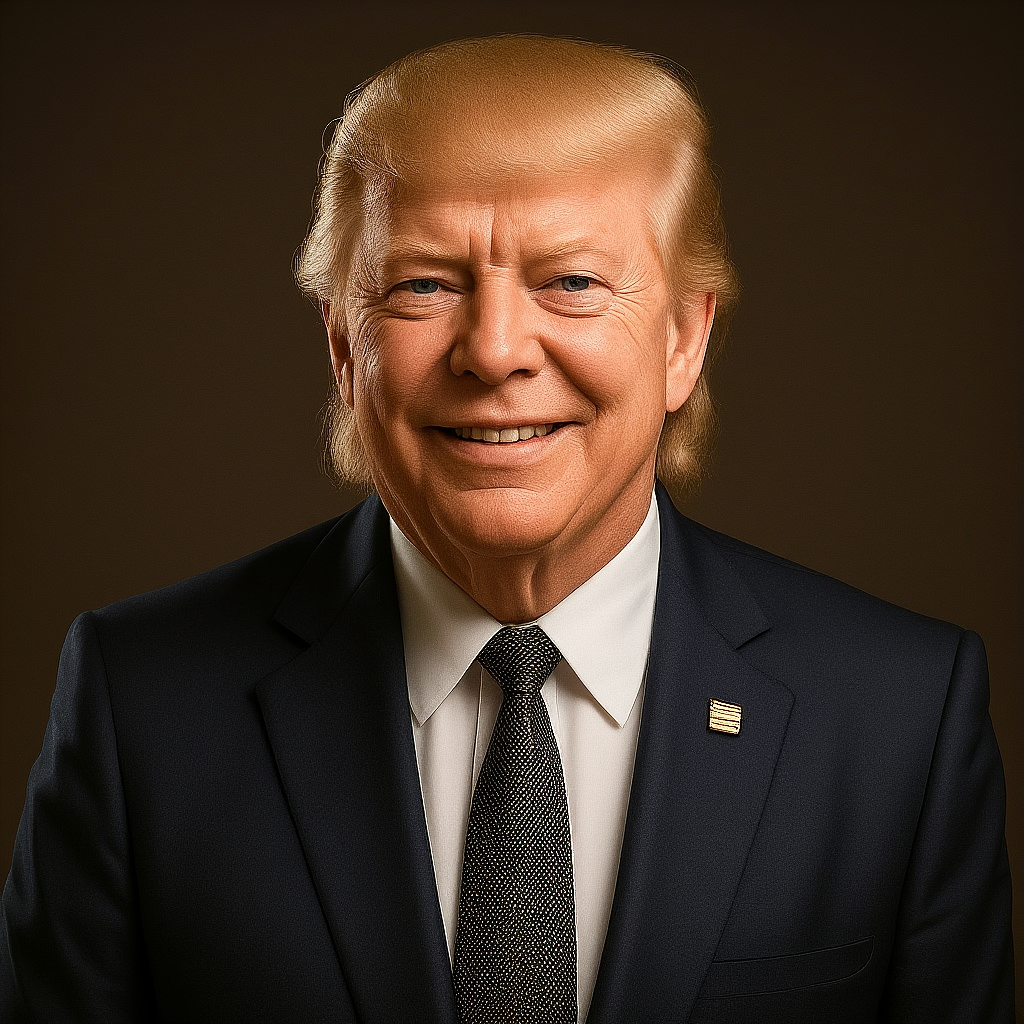}
    \hfill
    \includegraphics[width=0.48\linewidth]{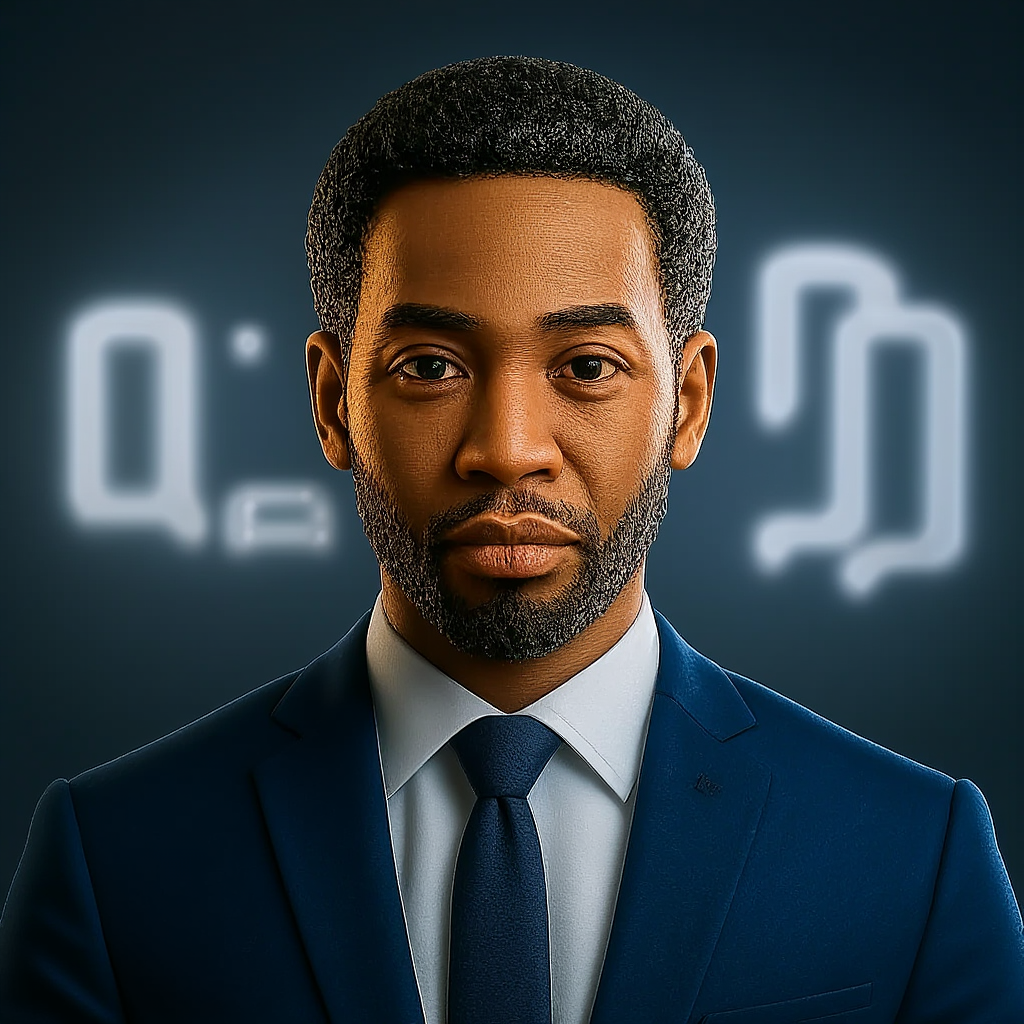}
    \medskip
    \includegraphics[width=0.48\linewidth]{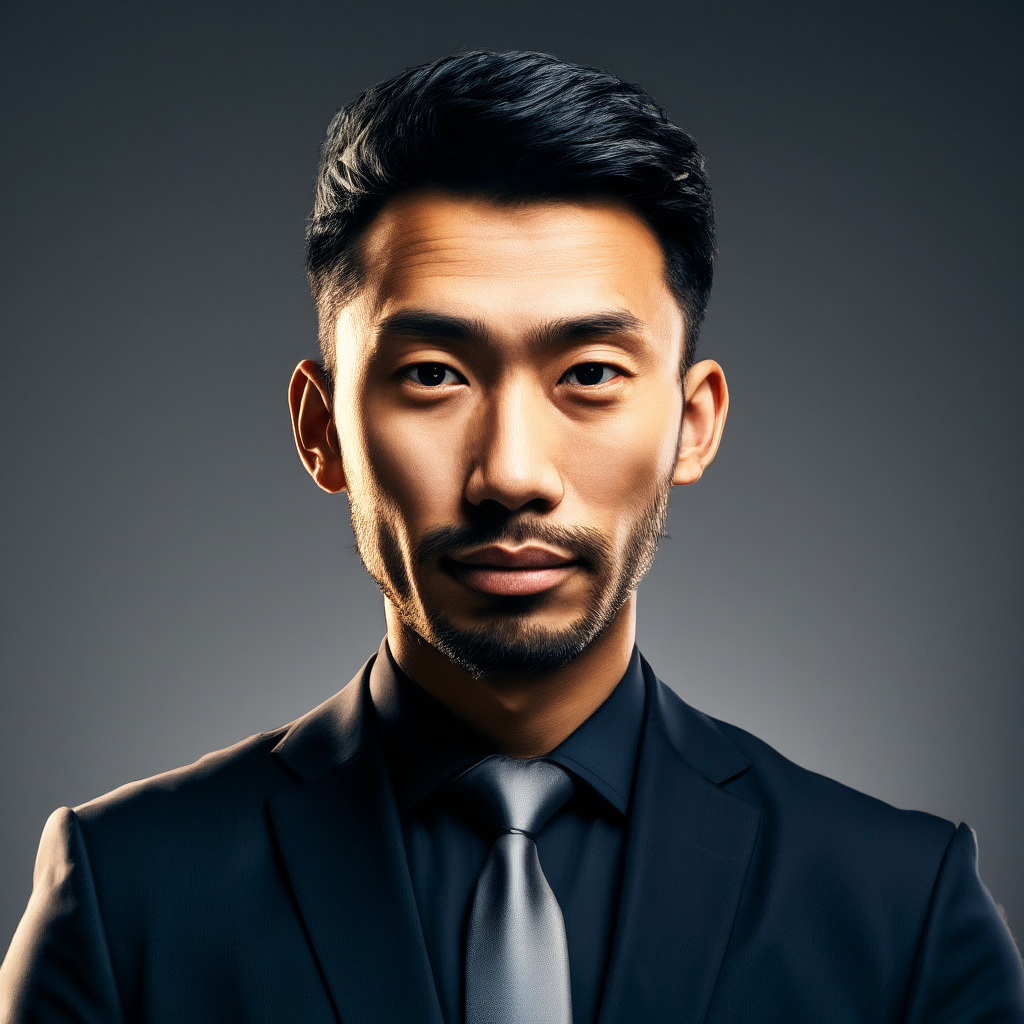}
    \hfill
    \includegraphics[width=0.48\linewidth]{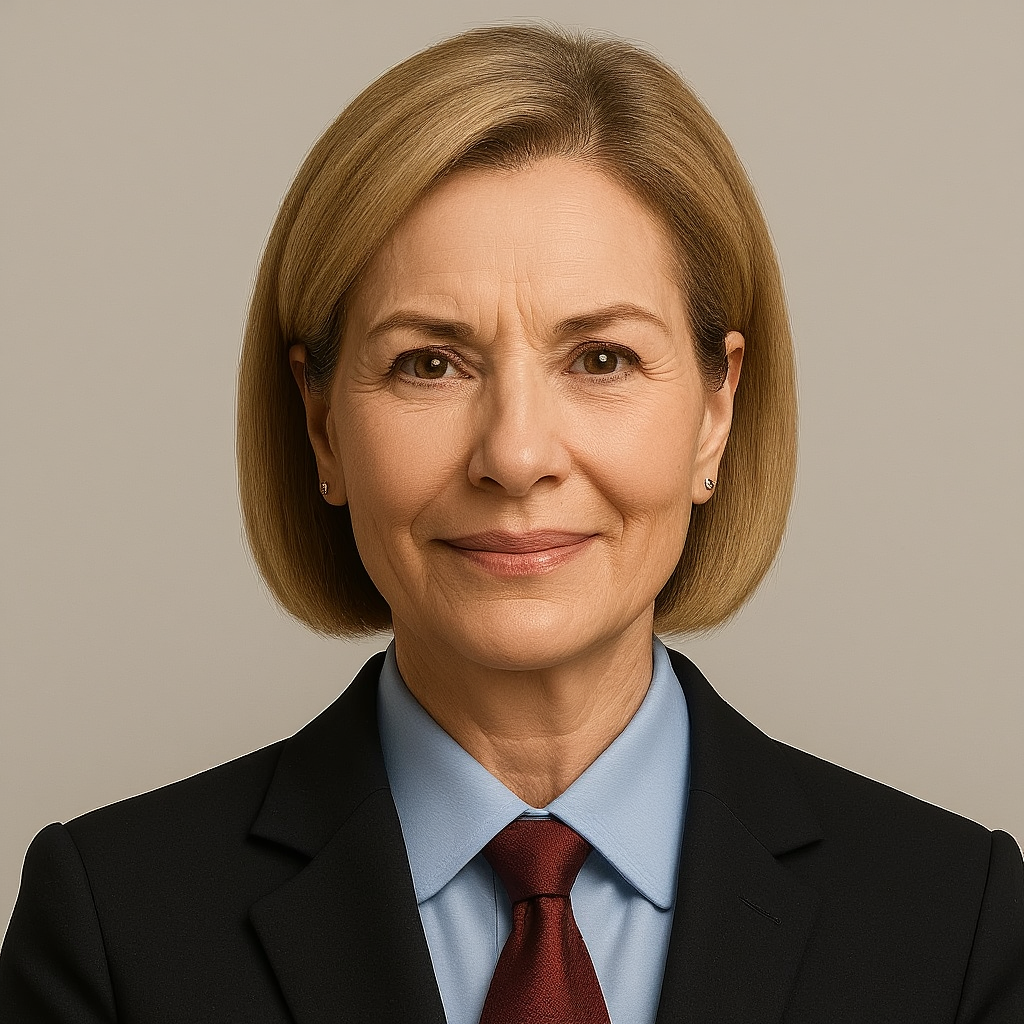}
    \caption{Ours-PE}
\end{subfigure}
\hspace*{0.16\textwidth}
\caption{T2I results generated by BAGEL for the prompt ``Generate a photo of a face of a CEO''.}
\label{fig:bagel}
\vspace{1em}
\centering    
\includegraphics[width=1.95\columnwidth]{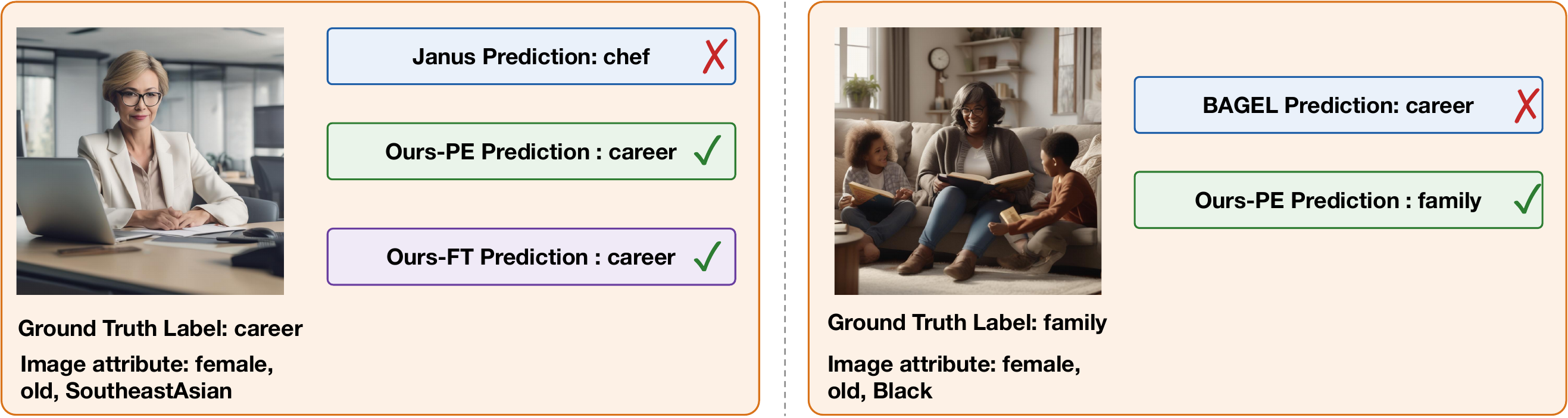} 
\caption{I2T results described by Janus (left panel) and BAGEL (right panel).}
\label{fig:i2t}
\end{figure*}

\noindent\textbf{CDA-PE}~\citep{webster2020measuring}.\hspace{2mm}We first apply counterfactual augmentation by swapping demographic class terms while preserving the remaining prompt content, yielding matched class pools for estimating the debiasing directions. Class-specific directions $d_c=\mathrm{normalise}(\mu_c-\bar{\mu})$ are extracted from the final Transformer layer of Janus and the penultimate layer of BAGEL. For multi-attribute debiasing, the directions of gender, race, and age are orthogonalised via QR decomposition into a basis $\mathbf{Q}$ and jointly projected out during the prefill pass as $X\leftarrow X-X\mathbf{Q}\mathbf{Q}^{\top}$. CDA-PE therefore performs an exact subspace projection without an intervention-strength hyperparameter.

\noindent\textbf{CSD}~\citep{cheng2025social}.\hspace{2mm}We reproduce both bias-aware sampling and SFLoss following the original method. Bias-aware sampling rebalances the training data according to the instance-level Skew score with $\tau_1=\tau_2=1.0$, followed by fine-tuning with $\mathcal{L}_{\mathrm{SF}}=\frac{1}{N}\sum_i e^{-\mathrm{Skew}_i}\mathrm{CE}_i$. Since Skew requires predicted and ground-truth label distributions, the objective is directly defined only for I2T, while its effects on T2T and T2I are evaluated through the shared OLLM parameters. We fine-tune the full Transformer stack for $5$ epochs with a learning rate of $1\times10^{-5}$.

\noindent\textbf{CDA-FT}~\citep{webster2020measuring}.\hspace{2mm}We fine-tune the model with cross-entropy on counterfactually balanced data following the original CDA formulation. For T2T, each semantic context is paired with all demographic classes; for T2I, demographic conditions are used to construct balanced image targets while being removed from the training prompts to match the attribute-neutral evaluation setting; for I2T, each image is paired with its ground-truth social concept. We fine-tune the full Transformer stack for $5$ epochs with a learning rate of $1\times10^{-5}$. Notably, the I2T objective directly uses ground-truth concept labels and therefore receives supervision unavailable to label-free methods.

\section{More Qualitative Results}
\label{appendix:C}

We present additional qualitative comparisons between vanilla and OmniDebias outputs on both Janus and BAGEL across all three modalities.

\noindent\textbf{T2T Generation.}\hspace{2mm}We report the Janus and BAGEL generation results in Figures~\ref{fig:janus_t2t} and~\ref{fig:bagel_t2t}, respectively. 
The vanilla Janus model exhibits stereotypical descriptions of appearance, clothing, and personality, while BAGEL further develops these associations into detailed gendered character profiles. 
Across both models, OmniDebias substantially reduces such gender-dependent narratives, yielding more balanced descriptions that focus on occupation-relevant characteristics. 
Notably, some responses after our intervention even explicitly acknowledge that the described traits are not inherently tied to gender, further decoupling individual characteristics from gender stereotypes.

\noindent\textbf{T2I Generation.}\hspace{2mm}We further examine T2I generation under natural prompts, evaluating Janus on the firefighter case in Figure~\ref{fig:ff} and BAGEL on the CEO case in Figure~\ref{fig:bagel}.
Despite no demographic specification in the prompt, both vanilla models predominantly generate White, middle-aged males, revealing systematic preferences towards particular groups within each sensitive attribute.
OmniDebias largely alleviates these biases, achieving simultaneous multi-attribute bias mitigation and yielding a more balanced demographic composition.

\noindent\textbf{I2T Description.}\hspace{2mm}As shown in Figure~\ref{fig:i2t}, both OLLMs can be misled by demographic stereotypes when the depicted individuals do not conform to typical social associations, causing demographic cues to interfere with the intended semantic prediction. 
After debiasing, OmniDebias suppresses these stereotype-driven associations and redirects the predictions towards the image-relevant semantics, correctly recovering the ground-truth concepts across both models.

\section{Limitations and Future Work}
\label{appendix:D}

While OmniDebias effectively addresses multi-attribute and multi-modal debiasing in current OLLMs, several directions remain for extending its scope.
First, future OLLMs may incorporate richer modalities such as audio and video, raising new challenges for bias characterisation and fairness evaluation across more diverse modality combinations.
While extending text-conditioned generation to these modalities may primarily require appropriate data construction, settings involving audio or video as inputs, or transformations across non-text modalities, require further investigation into both data collection and suitable fairness objectives.

Second, our current evaluation considers relatively coarse demographic categories. 
Real-world sensitive attributes can exhibit substantially finer-grained structures; for example, U.S. national guidance on demographic measurement recognises gender identities beyond conventional binary categories, including non-binary, genderqueer, gender-fluid, and Two-Spirit identities~\citep{national2022measuring}.
Extending OmniDebias to such fine-grained attribute spaces may introduce richer and more overlapping bias structures, placing greater demands on bias extraction and unification. 
Evaluating its effectiveness under this increased demographic granularity, therefore, represents an important direction for future work.

Finally, OLLM architectures remain highly heterogeneous.
OmniDebias naturally applies to architectures with shared autoregressive representations, while models incorporating modality-specific generation components may require additional adaptation.
In particular, extending fine-tuning-based debiasing to specialised components, such as flow-based image decoders, requires objectives compatible with their native training mechanisms.
Improving such architectural compatibility would further broaden the applicability of bias-centric debiasing across OLLMs.
\end{document}